\documentclass[sn-mathphys-num,,Numbered]{sn-jnl}
\usepackage{graphicx}
\graphicspath{ {Images/} }
\usepackage[caption=false]{subfig} 
\usepackage{float}
\usepackage{enumitem}

\usepackage{amsmath,amssymb,amsfonts} 

\usepackage{hyperref}
\hypersetup{colorlinks=true, citecolor=blue, urlcolor=black, linkcolor=magenta}
\usepackage{seqsplit}
\usepackage{xurl}
\usepackage{bbding} 

\usepackage{xltabular}
\usepackage{tabularx, tabularray, multirow}
\usepackage{array}
\newcolumntype{C}{>{\centering\arraybackslash}X}
\newcolumntype{L}{>{\raggedright\arraybackslash}X}
\newcolumntype{R}{>{\raggedright\arraybackslash}X}

\usepackage{comment} 
\usepackage{soul}
\usepackage{microtype}
\newcommand{\Revise}[1]{\textcolor{black}{#1}}

\begin{document}\microtypesetup{activate=true}

\title{An Investigation of Visual Foundation Models Robustness}
\author*[1,2]{\fnm{Sandeep} \sur{Gupta}}\email{s.gupta@qub.ac.uk}
\author[2]{\fnm{Roberto} \sur{Passerone}}\email{roberto.passerone@unitn.it}

\affil*[1]{\orgdiv{Centre for Secure Information Technologies (CSIT)}, \orgname{Queen's University Belfast}, \orgaddress{\country{UK}}}
\affil[2]{\orgdiv{Department of Information Engineering and Computer Science}, \orgname{University of Trento}, \orgaddress{\country{Italy}}}

\abstract{Visual Foundation Models (VFMs) are becoming ubiquitous in computer vision, powering systems for diverse tasks such as object detection, image classification, segmentation, pose estimation, and motion tracking. VFMs are capitalizing on seminal innovations in deep learning models, such as LeNet-5, AlexNet, ResNet, VGGNet, InceptionNet, DenseNet, YOLO, and ViT, to deliver superior performance across a range of critical computer vision applications. These include security-sensitive domains like biometric verification, autonomous vehicle perception, and medical image analysis, where robustness is essential to fostering trust between technology and the end-users. This article investigates network robustness requirements crucial in computer vision systems to adapt effectively to dynamic environments influenced by factors such as lighting, weather conditions, and sensor characteristics. We examine the prevalent empirical defenses and robust training employed to enhance vision network robustness against real-world challenges such as distributional shifts, noisy and spatially distorted inputs, and adversarial attacks. Subsequently, we provide a comprehensive analysis of the challenges associated with these defense mechanisms, including network properties and components to guide ablation studies and benchmarking metrics to evaluate network robustness.
}

\maketitle

\section{Introduction}
\begin{figure*}[!ht]
    \centering
    \includegraphics[width=1\linewidth, keepaspectratio]{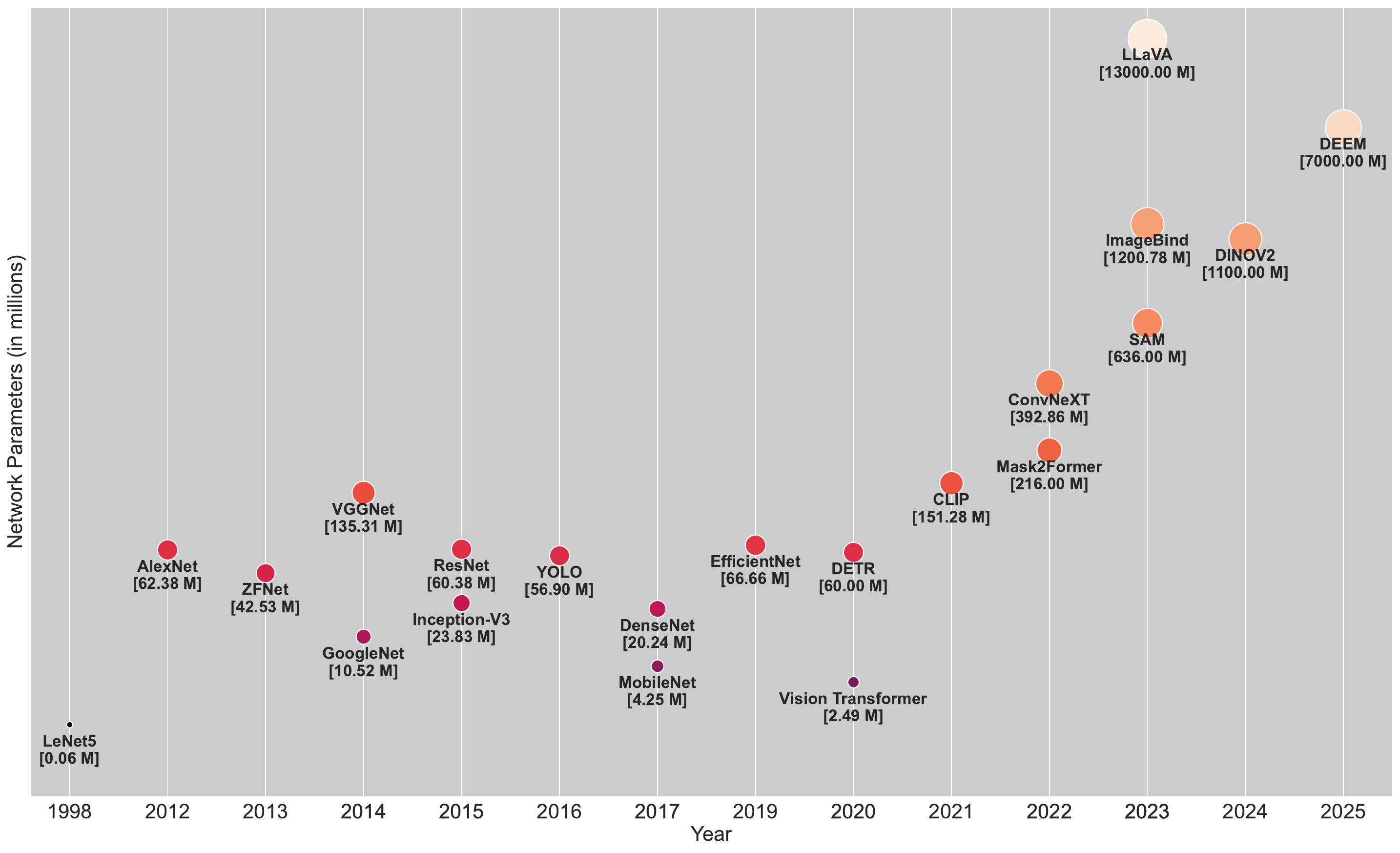}
    \caption{A timeline providing a glimpse of landmark network architectures, transformers, encoders, and foundation models in computer vision. LeNet-5~\cite{lecun1998gradient} design for handwritten digit recognition became the foundation for later Convolutional Neural Networks (CNNs). AlexNet~\cite{krizhevsky2012imagenet} architecture optimizes training for large-scale image datasets. ZFNet~\cite{zeiler2014visualizing} introduces visualization for debugging and ablation study. VGGNet~\cite{simonyan2015vggnet} uses uniform architecture with small 3x3 convolutional kernels and deep stacks of layers. GoogleNet~\cite{szegedy2015going} introduces the Inception module enabling parallel convolutional operations with different kernel sizes. ResNet~\cite{ren2016deep} architecture addresses the vanishing gradient problem with an introduction of residual or skip layers. Inception-V3~\cite{szegedy2016rethinking} architecture introduces label smoothing, factorized convolutions, and the use of an auxiliary classifier. YOLO~\cite{redmon2016you} architecture supports single-stage detection by combining region proposal algorithms with CNNs. DenseNet~\cite{huang2017densely} features dense connections between layers. MobileNet~\cite{howard2019searching} facilitates efficient deployment on mobile and edge devices. EfficientNet~\cite{tan2019efficientnet} features uniform scaling of all dimensions (i.e., depth, width, resolution) using compound coefficient. ViT~\cite{dosovitskiy2021an} leverages the Transformer architecture enabling self-attention across image patches. DEtection TRansformer (DETR)~\cite{carion2020end} is end-to-end object detection based on transformers and ResNet framework with bipartite matching loss approach. CLIP~\cite{radford2021learning} model is based on the Vision Transformer architecture (ViT-B/32) and ResNet-50, enabling it to generalize to arbitrary image classification tasks in a zero-shot manner. Mask2Former~\cite{cheng2022masked} is masked-attention mask transformer for universal image segmentation supporting panoptic, instance and semantic segmentation. ConvNeXts~\cite{liu2022convnet} demonstrate competitive performance against state-of-the-art hierarchical vision Transformers across multiple computer vision benchmarks, despite their simpler ConvNet-based architecture. ImageBind~\cite{girdhar2023imagebind} provides a joint embedding for six different modalities: images, text, audio, depth, thermal, and IMU data  for zero-shot classification. Segment Anything Model (SAM)~\cite{kirillov2023segment} facilitates promptable visual segmentation in images and videos. LLaVA~\cite{liu2023visual} is a large multimodal model (LMM) that combines a vision encoder with Vicuna for general-purpose visual and language understanding. DINOV2~\cite{oquab2024dinov} trained with different Vision Transformers on large curated data with no supervision. DEEM~\cite{luo2025deem} enhances the semantic alignment of its image encoder by leveraging the generative feedback mechanism inherent in diffusion models.}\label{fig:FoundationModels}
\end{figure*}

Visual Foundation Models (VFMs) represent a significant advancement in computer vision, demonstrating versatility through superior performance across a range of tasks, including object detection, image classification, segmentation, pose estimation, and motion tracking~\cite{gupta2024visual}. VFMs are pre-trained deep neural networks (DNNs) that leverage massive and diverse datasets to learn universal visual features, making them applicable to various downstream tasks across multiple domains~\cite{liu2024few}. They offer multidimensional capabilities, serving both as embedding extractors for transferring knowledge from pre-trained models to specific tasks, and as zero- or one-shot learners capable of directly executing tasks on unseen data. Figure~\ref{fig:FoundationModels} provides a timeline of several landmark DNN architectures, highlighting advancements in computer vision since LeNet-5~\cite{lecun1998gradient} was developed to recognize handwritten and machine-printed digits. Substantial performance improvements of DNNs are evident in the progression from AlexNet~\cite{krizhevsky2012imagenet}, VGGNet~\cite{simonyan2015vggnet}, and ResNet~\cite{ren2016deep} to InceptionNet~\cite{szegedy2016rethinking}, DenseNet~\cite{huang2017densely}, You only Look Once (YOLO)~\cite{redmon2016you}, and Vision Transformer (ViT)~\cite{dosovitskiy2021an}. Transformer architecture excels at processing data sequences through attention mechanisms that dynamically evaluate the importance of different inputs. Recent networks such as Contrastive Language-Image Pre-training (CLIP)~\cite{radford2021learning}, ImageBind~\cite{girdhar2023imagebind}, Large Language and Vision Assistant (LLaVA)~\cite{liu2023visual}, Deeper
Into Neural Networks (DINOv2)~\cite{oquab2024dinov}, and Diffusion Models Serve as the Eyes of Large Language Models (DEEM)~\cite{luo2025deem} have demonstrated efficiency in multi-modal data understanding, generating richer embeddings, supporting both visual and non-visual tasks, and advancing self-supervised learning (SSL). These architectures optimize training for large-scale image datasets, support parallel convolutional operations with different kernel sizes, address the vanishing gradient problem through the introduction of residual or skip layers, single-stage detection by combining region and self-attention mechanisms across image patches.

VFMs can drive transformative advancements in security-critical domains such as biometric verification, autonomous vehicle perception, and medical image analysis. Despite the remarkable performance of VFMs in computer vision, they can be vulnerable to distributional shifts~\cite{gupta2025evaluating,shu2021encoding}, noisy and spatially distorted inputs~\cite{zhu2023understanding}, and adversarial perturbation~\cite{papernot2016distillation}. Since robustness is critical to establishing trust between technology and end-users, addressing these challenges is crucial for enhancing model robustness in real-world scenarios. 
For instance, understanding distributional shifts, such as semantic and covariate shifts, can be useful in mitigating the impact of natural corruptions~\cite{gardner2024benchmarking}. This is especially relevant in medical imaging, where datasets inherently vary due to diverse patient populations, imaging protocols, and equipment differences. Similarly, robustness of models to noisy and spatially distorted inputs is critical for real-world applications, such as biometric systems, and connected and autonomous vehicles, where sensor malfunctions, object appearance variations, or environmental factors often introduce such distortions~\cite{feng2023robust}. Furthermore, adversarial robustness is essential for defending against digitally generated adversarial examples, which adversaries craft with either full (white-box) or limited/no (gray/black-box) access to the target model~\cite{bitton2023evaluating, meyers2023safety}. 

Network robustness\footnote{In this paper, we use `network or model robustness' as a collective term for the robustness of a VFM, Vision transformer, or image encoders in VLMs models.} is ideally defined by the ability to generalize to unseen data and maintain performance despite transformations, noise, errors, and adversarial attacks. The emphasis on network robustness not only enhances reliability but also broadens the applicability of FMs to diverse and sensitive domains. Our investigation is guided by the following research questions:
\begin{enumerate}[leftmargin=*,label=\textbf{RQ\arabic*.},itemsep=0pt]
    \item What are the primary requirements for network robustness in the context of computer vision?
    \item What are the prevalent mechanisms employed to improve network robustness?
    \item What challenges are associated with defense mechanisms employed to enhance network robustness?    
    \item What aspects, i.e., the properties and components, of a network should be analyzed for strengthening network robustness?
    \item What quantitative metrics can be used to benchmark network robustness?
\end{enumerate}
\vspace{4pt}

This article elucidates the requirements for network robustness in computer vision, discussing distributional shifts, noisy or spatially distorted inputs, and adversarial perturbations, which can enable reliable and widespread real-world deployment. We analyze the empirical defenses and robust training mechanisms used to improve network robustness. Subsequently, we compare various defense mechanisms, including adversarial detection, input transformations, adversarial training, certified defenses, network distillation, and adversarial learning. We further highlight network features and components suitable for ablation studies, along with relevant benchmarking metrics for assessing network robustness. The rest of the paper is organized as follows: Section~\ref{sec:Background} provides an overview of visual foundation models, the definition of network robustness, and the notation used throughout the paper. Section~\ref{sec:RobustnessReq} outlines the primary requirements for network robustness, addressing \textbf{RQ1}. Section~\ref{sec:Defense} discusses potential defense mechanisms to enhance network robustness in response to \textbf{RQ2}. Section~\ref{sec:Discussions} examines \textbf{RQ3}, \textbf{RQ4}, and \textbf{RQ5}, focusing on advancing the state-of-the-art in robust network training and empirical defense methods. Finally, Section~\ref{sec:Conclusions} concludes the article.

\section{Background}\label{sec:Background}
This section provides a brief overview of visual foundation models and network robustness. The notations used throughout the paper are listed below.
\begin{table*}[!ht]
    \centering
    \footnotesize
    \hyphenpenalty 10000
    \begin{tblr}{
        width=1\linewidth,
        colspec = {p{.18\linewidth}  p{.75\linewidth}},
    }\hline
    \textbf{Notation} & \textbf{Description} \\\hline
    $f(\cdot)$ & A neural network.\\\hline
    $f^{1...j}(\cdot)$ & The composition of the neural network from layers $1$ to $j$. \\\hline
    $f^j(\cdot)$ &  $j^{th}$ layer of the neural network.\\\hline
    $x$ & Input image (Benign/clean sample without any malicious perturbation).\\\hline
    $y$ & Input image label (Ground truth).\\\hline
    $f(x)_y$ & Probability that the image $x$ corresponds to the label $y$. \\\hline
    $y_{pred}$ & The label predicted by the network obtained using $\underset{y}{\arg\max}~f(x)_y$. \\\hline
    $x'$ & Adversarial image (Sample with any malicious perturbation).\\\hline
    $\epsilon$ & Magnitude of the perturbation.\\\hline
    $\ell_{p}~\vert~p~\in 0,1,2,\infty$ & Perturbation metric to quantify the strength and size of distortions made on input data\\\hline
    \end{tblr}
\end{table*}

\subsection{Visual foundation models}\label{sec:VFMs}
Visual foundation models are pre-trained models that are typically trained on millions of image-text pairs. VFM training data generation can be automated through self-supervised data curation using models such as CLIP and DINOv2~\cite{xu2024demystifying}. CLIP learns a multi-modal embedding space by jointly training an image- and text encoder, whereas DINOv2 can learn from any collection of images. SSL supports tasks like segmentation, classification, and regression, offering a scalable alternative to manual data labeling, which is both costly and time-consuming. Additionally, large language models can facilitate image generation from text prompts, further expanding the training data pool. As a result, VFMs effectively learn visual features, generating representations with strong generalization and transfer capabilities. These representations embed both image and text features into a shared semantic space, enabling their application to a wide range of downstream tasks in a few-shot manner~\cite{liu2024few}. The ability of VFMs to generate coherent and contextually relevant outputs for image-text pairs is largely driven by the integration of ViT architectures~\cite{liu2022convnet}. Unlike CNNs~\cite{li2021survey}, which perform convolutional operations on local image patches to extract features such as objects, edges, and textures, ViTs~\cite{dosovitskiy2021an} employ a transformer-based architecture that converts the entire image into a sequence of vectorized embeddings, thereby capturing long-range dependencies. Critically, the self-attention mechanism within the transformer enables the model to capture relationships between different parts of an image, leading to a richer understanding of visual content.

Vision-language models (VLMs), such as LLaVA, and DEEM, integrate image encoders with a large language model (LLM), enabling them to process and interpret video, image, and text prompts to generate insightful textual responses. Table~\ref{tab:NetworksSummary} presents an overview of recent VFMs and VLMs. For instance, LLaVA is a general image-to-text generative model framework that consists of two pre-trained modules, i.e., a vision encoder (e.g., CLIP) and a LLM (e.g., Vicuna), making it suitable as a general-purpose visual assistant~\cite{liu2023visual}. DEEM leverages the generative feedback of diffusion models to align the semantic distributions of the image encoder in a self-supervised manner~\cite{luo2025deem}. DEEM leverages ConvNeXt as the image encoder, and Vicuna and Stable Diffusion v2.1 as the language model and image decoder, respectively. VLMs process interleaved image-text pairs by first encoding them into visual and textual tokens using dedicated encoders. These tokens, arranged according to their original layout, are then fed into a large language model to produce hidden state outputs. Leveraging auto-regressive modeling for the text hidden states, the model uses the image output hidden states and the encoded image tokens as diffusion conditions. These conditions guide a diffusion model to reconstruct the input image. Through end-to-end training, VLMs learn to generate both text and images and employ semantic consistency regularization on the image encoder's semantic output during reconstruction. VLMs are powerful tools for tasks like text-to-image generation, image-grounded text generation (e.g., image captioning, visual question answering), and joint generation~\cite{zhao2023evaluating}. Thus, visual robustness is a prerequisite for preventing visual hallucinations and ensuring reliable visual perception in VLMs. 

\begin{table*}[!ht]
\Revise{
    \centering
    \footnotesize
    \hyphenpenalty 10000
    \caption{An overview of recent VFMs and VLMs}\label{tab:NetworksSummary}
    \begin{tblr}{
        width=1\linewidth,
        colspec = {p{.15\linewidth} p{.03\linewidth}  p{.22\linewidth} p{.22\linewidth} p{.22\linewidth}},
    }\hline
    \textbf{Network} & \textbf{Year} & \textbf{Image Encoder} & \textbf{Text Encoder} & \textbf{Usage}\\\hline
    DEEM~\cite{luo2025deem} & 2025 & ConvNeXT & Vicuna & To enhance foundational visual perception capabilities\\\hline
    DINOV2~\cite{oquab2024dinov} & 2024 & ViT & - & Learning robust visual features without supervision\\\hline
    LLaVA~\cite{liu2023visual} & 2023 & CLIP & Vicuna, LLaMA & Language and vision assistant\\\hline
    ImageBind~\cite{girdhar2023imagebind} & 2023 & ViT & Design inspired from CLIP & A joint embedding space for modalities like image/video, text, audio, depth, IMU and thermal images\\\hline
    SAM~\cite{kirillov2023segment} & 2023 & ViT & - & Visual segmentation in images and videos\\\hline
    Mask2Former~\cite{cheng2022masked} & 2022 & ResNet-50, ResNet-101, SWIM & - & Panoptic, instance and semantic segmentation\\\hline
    CLIP~\cite{radford2021learning} & 2021 & ResNet-50, ViT-B/32 & A masked self-attention Transformer & Visual classification benchmark\\\hline
    DETR~\cite{carion2020end} & 2020 & ResNet-50, ResNet-101 & - & Objection detection\\\hline
    \end{tblr}
}
\end{table*}

\subsection{VFMs architecture}
\begin{figure*}[!ht]
    \captionsetup{format=hang,font=small, margin=5pt}
    \hyphenpenalty 10000
    \centering
     \subfloat[Encoder-decoder architecture features a single encoder paired with either a single decoder~\cite{badrinarayanan2017segnet} or multiple decoders~\cite{usman2024enhanced}.\label{fig:VFMsArch_DE}]{
        \includegraphics[width=0.44\linewidth, keepaspectratio]{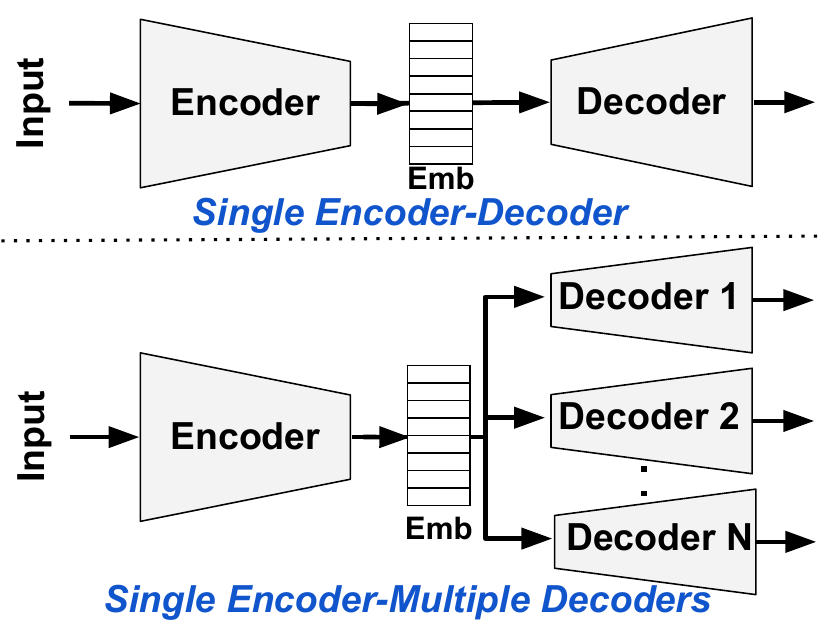}
    }
    \subfloat[Dual-encoder architecture features two encoders designed to handle heterogeneous inputs~\cite{dong2022exploring}.\label{fig:VFMsArch_Dual}]{
        \includegraphics[width=0.4\linewidth, keepaspectratio]{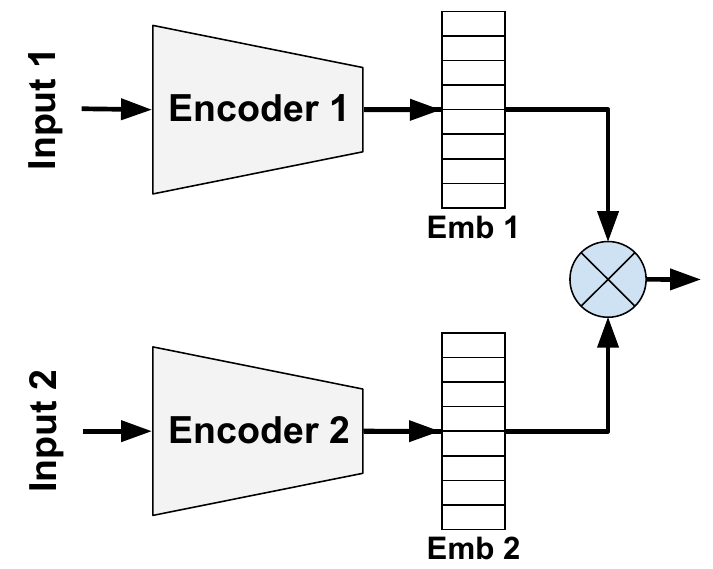}
    }
   \\
    \subfloat[Fusion architecture~\cite{tziafas2023early} employs either early or late fusion strategies to combine encoder outputs into a unified representation, tackling complex cross-modal interactions.\label{fig:VFMsArch_Fusion}]{
        \includegraphics[width=0.9\linewidth, keepaspectratio]{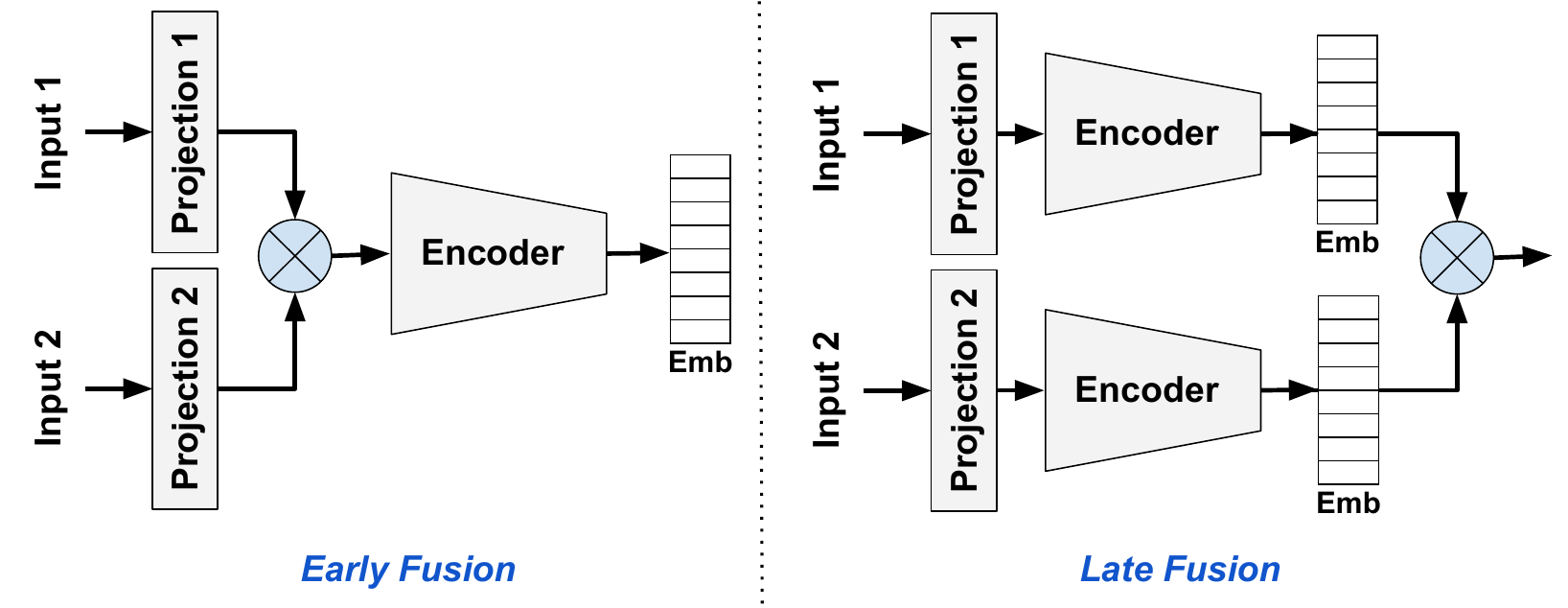}
    }
    \caption{Illustration of various architectures use for designing visual foundation models}\label{fig:VFMsArch}
\end{figure*}

Figure~\ref{fig:VFMsArch} illustrates the design diversity of VFMs, encompassing encoder-decoder~\cite{badrinarayanan2017segnet,usman2024enhanced}, dual-encoder~\cite{gao2021simcse}, and fusion~\cite{tziafas2023early,huang2024early} architectures, to efficiently process visual content for downstream tasks such as object detection, image classification, segmentation, pose estimation, and motion tracking. Expanding the capabilities to understand both visual and textual information, VLMs are typically built on the foundation of a pre-trained visual encoder and a pre-trained LLM with cross-modal understanding capabilities.

\begin{itemize}[leftmargin=*]
\setlength\itemsep{1em}
    \item The \textit{encoder-decoder architecture} features a single encoder paired with either a single decoder~\cite{badrinarayanan2017segnet} or multiple decoders~\cite{usman2024enhanced}, enabling it to address a single task or multiple related tasks respectively, as shown in Figure~\ref{fig:VFMsArch_DE}. An encoder compresses input data into a lower-dimensional representation, and a decoder generates output data from this representation by mapping the low-resolution feature maps to the full input resolution.

    \item The \textit{dual-encoder architecture} employs two distinct encoders to process input data pairs, such as text and images~\cite{gao2021simcse}. Subsequently, the encoded representations are fused, often using a contrastive loss function to synchronize the embeddings from the encoders, to produce the target output, as shown in Figure~\ref{fig:VFMsArch_Dual}.
    
    \item The \textit{fusion architecture} combines individual encoder outputs into a unified representation, facilitating complex cross-modal interactions~\cite{huang2024early}. As shown in Figure~\ref{fig:VFMsArch_Fusion}, these architectures commonly employ early or late fusion strategies~\cite{tziafas2023early}. In the early-fusion approach, all inputs are projected into a shared representational space from the start, enabling seamless reasoning and generation across inputs. In contrast, the late-fusion approach maintains separate encoders for each input, encoding images and text separately before combining them at a later stage.
\end{itemize}

\subsection{Network robustness}
Network robustness refers to a network ability to maintain reliable, stable, and consistent performance for specified inputs in empirical environments~\cite{zhu2023understanding}. Ideally, the robustness of a network can be specified as its ability to generalize to new and unseen data and be resistant to transformations, noise, errors, and adversarial attacks~\cite{li2023trustworthy}. In a broader context, the robustness of an AI system can be assessed by its ability to resist attacks that could compromise its confidentiality, integrity, or availability~\cite{gupta2024visual}.

Bhojanapalli et al.~\cite{bhojanapalli2021understanding} investigate robustness to input perturbations and model perturbations. The authors highlight that evaluating the robustness of a network against input perturbations requires benchmarking on naturally occurring image corruptions (e.g., noise and blur), spatial perturbations (e.g., translations and rotations), adversarial perturbations, and texture bias. Robustness to model perturbations requires computing layer correlations, conducting lesion studies, and focusing attention selectively. Gu et al.~\cite{gu2022evaluating} identify two primary types of patch perturbations for evaluating network robustness. The first type involves natural corruptions arising from distributional shifts, while the second consists of adversarial perturbations designed to intentionally mislead network predictions. Zhao et al.~\cite{zhao2023evaluating} evaluate the adversarial robustness of VLMs by crafting targeted adversarial examples using pretrained CLIP models as surrogates, by matching either textual or image embeddings. Subsequently, these adversarial examples are transferred to other large VLMs, where these transfer- and query-based attacks achieve a high success rate in inducing targeted responses.

Studies~\cite{han2023interpreting, qian2022survey} report that adversarial patches are among the most widely used methods for physical adversarial attacks on computer vision models, where printed patches and stickers are applied to objects to induce misclassification. Hendrycks and Dietterich~\cite{hendrycks2018benchmarking} introduce the IMAGENET-C and IMAGENET-P datasets for evaluating the robustness of CNNs to input corruptions and perturbations. The IMAGENET-C dataset includes 75 common visual corruptions related to noise, blur, weather, and digital distortions, while the IMAGENET-P dataset contains perturbation sequences generated from motion blur, zoom blur, snow, brightness, translation, rotation, tilt (viewpoint variation through minor 3D rotations), and scale perturbations. 
Meng et al.~\cite{meng2022adversarial} present a formal verification approach for adversarial robustness of deep neural networks, focusing on property formalization, reduction frameworks, and reasoning strategies. Qian et al.~\cite{qian2022survey} outline detection-based and denoising-based methods for establishing model robustness. Detection-based methods employ techniques to identify and classify input samples as either adversarial or benign, while denoising methods aim to eliminate adversarial perturbations prior to image classification.

\section{Robustness requirements}\label{sec:RobustnessReq}
In response to \textbf{RQ1} (\textit{What are the primary requirements for network robustness in the context of computer vision}?), this section discusses the primary requirements for network robustness. In safety-critical applications like biometric verification, autonomous driving, and medical image analysis, trust between technology and end-users hinges on the ability of computer vision systems to reliably adapt to dynamic environments. These environments are often characterized by fluctuating lighting, weather conditions, sensor variations and noisy inputs. It is therefore recommended to expand the scope of threat models beyond the adversary’s \textit{goals}, \textit{specificity}, and \textit{knowledge} to establish network robustness requirements.

\begin{itemize}[leftmargin=*]
\setlength\itemsep{1em}
\item \textbf{Goal:} In poisoning attacks~\cite{fan2022survey}, the adversary can access and modify the training dataset by injecting fake samples into the training data to generate a defective model. Thus, the final trained models can be less accurate and more misclassification. Whereas, in evasion attacks~\cite{wang2023evasion}, the adversary can attack a well-trained model with deceptive test samples such that the model fails to recognize them, thus, evading the final detection during the the inference phase.
    
\item \textbf{Specificity:} In untargeted attacks~\cite{li2024survey}, the adversary deceives a model such that the model predicts falsely irrespective of the label generated. Whereas, in targeted attacks~\cite{ren2020adversarial}, the adversary aims that a model predicts incorrectly for a specific label.

\item \textbf{Knowledge:} White-box attacks~\cite{chen2024learn} assume that the adversary possesses full access to the internal components of the target model, including its architectural design, gradient information, and training parameters. Gray-box attacks~\cite{serban2020adversarial} involve an adversary who lacks knowledge of the target model architecture. Instead, the adversary trains a surrogate model using the available training data and exploits the transferability of adversarial examples to mimic the target model behavior. 
Black-box attacks~\cite{zhao2023remix} refer to scenarios in which an adversary seeks to compromise a model without access to its internal architecture or parameters, instead leveraging limited information derived from the model outputs in response to specific input queries.
\end{itemize}

\begin{figure*}[!ht]
    \captionsetup{format=hang,font=small, margin=3pt}
    \hyphenpenalty 10000
    \centering
    \subfloat[Original scene is predicted as arterial road.]{
        \includegraphics[width=0.23\linewidth, keepaspectratio]{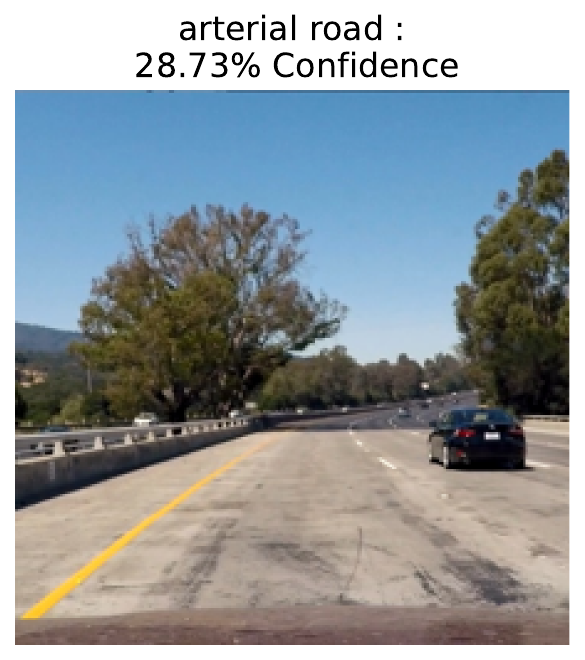}
    }
    \subfloat[Distributional shifts: snowy condition. Arterial road is predicted as tarmac.]{
        \includegraphics[width=0.23\linewidth, keepaspectratio]{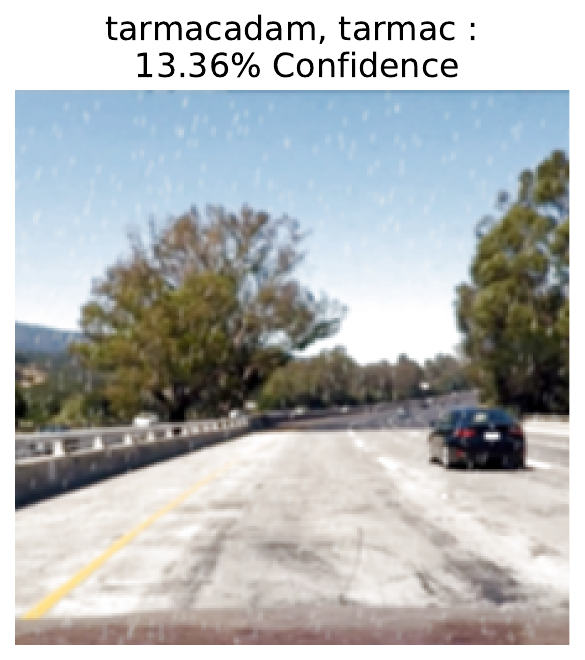}
    }
    \subfloat[Noisy and spatial translation: The scene is identified as blind curve or bend due to motion blur.]{
        \includegraphics[width=0.23\linewidth, keepaspectratio]{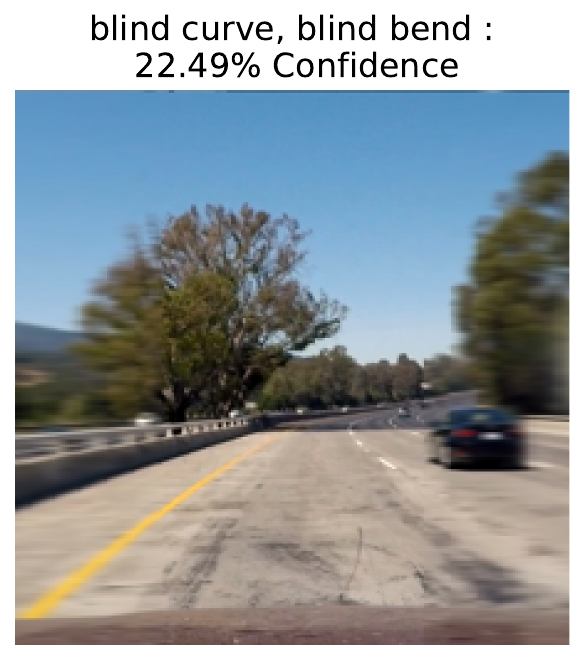}
    }
    \subfloat[Adversarial: PGD attack with $\epsilon=0.01$ the scene is identified as crash barrier.]{
        \includegraphics[width=0.23\linewidth, keepaspectratio]{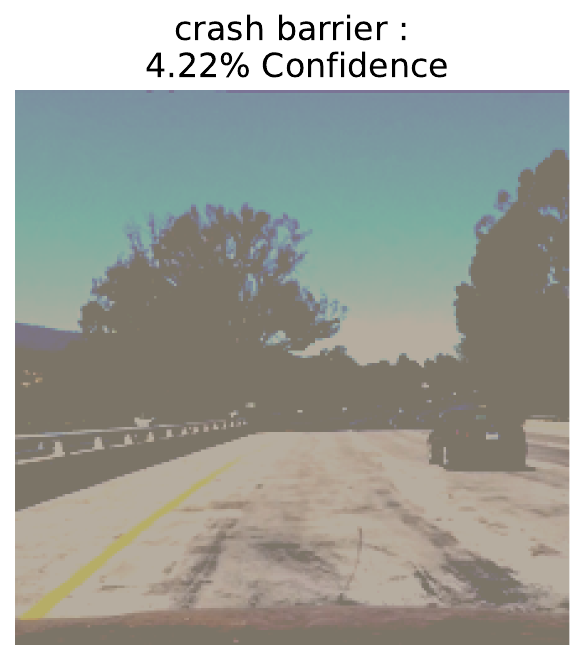}
    }
    \caption{Illustration of the prediction results of the ConvNeXT-XL model~\cite{liu2022convnet}, trained on the ImageNet22K dataset, simulating autonomous driving scenarios. ConvNeXT, a simplified ConvNet architecture, modernizes DNN designs by incorporating transformer-style self-attention and parallel processing.}\label{fig:AAConvNeXT}
\end{figure*}

\begin{figure*}[!ht]
\Revise{
    \captionsetup{format=hang,font=small, margin=2pt}
    \hyphenpenalty 10000
    \centering
    \subfloat[Vehicle is classified as golf cart.]{
        \includegraphics[width=0.23\linewidth, keepaspectratio]{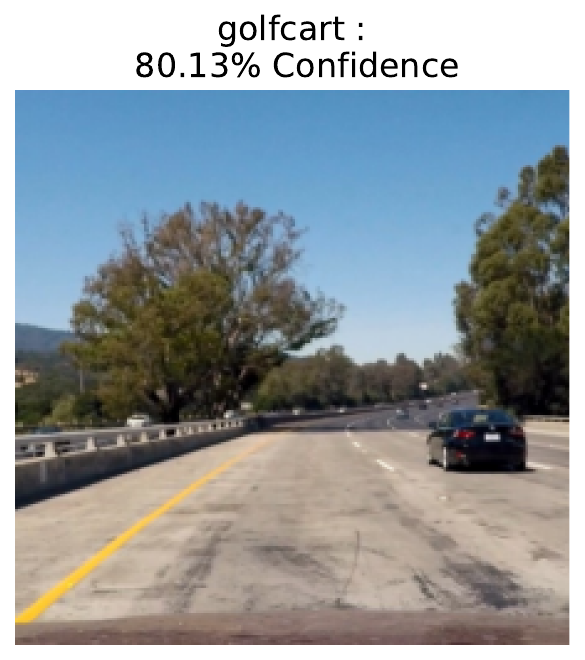}
    }
    \subfloat[Distributional shifts: snowy condition. Vehicle is predicted as go-kart.]{
        \includegraphics[width=0.23\linewidth, keepaspectratio]{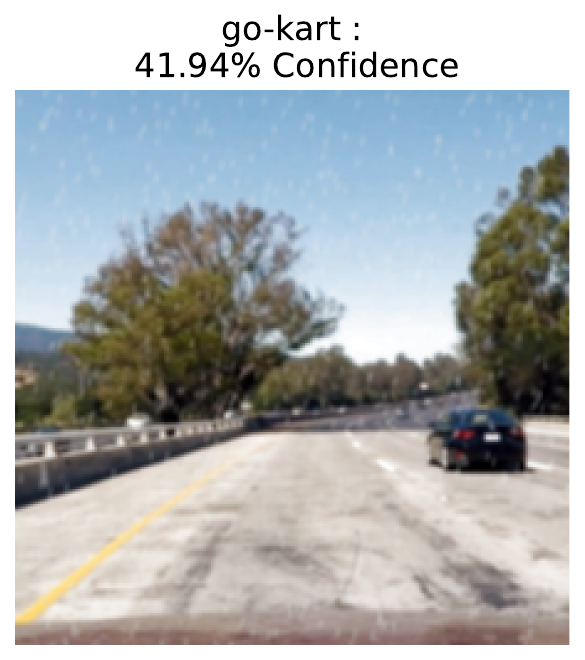}
    }
    \subfloat[Noisy and spatial translation: Lens glare. Vehicle is misclassified as planetarium.]{
        \includegraphics[width=0.23\linewidth, keepaspectratio]{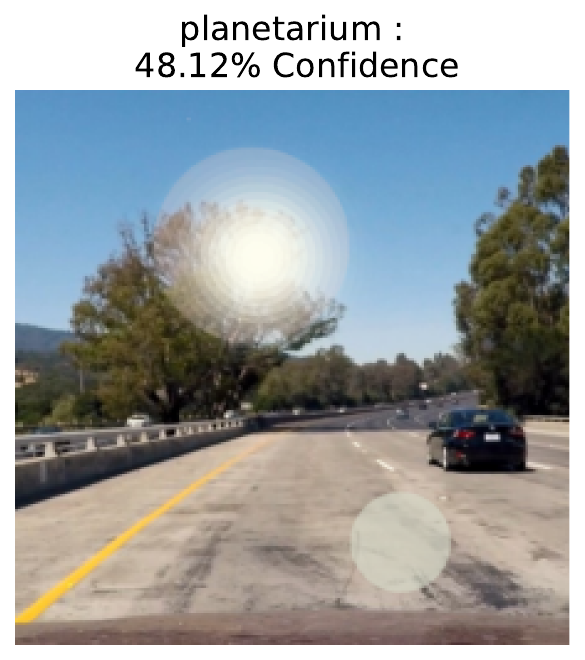}
    }
    \subfloat[Adversarial: PGD attack with $\epsilon=0.01$ the vehicle is identified as lumbermill.]{
        \includegraphics[width=0.23\linewidth, keepaspectratio]{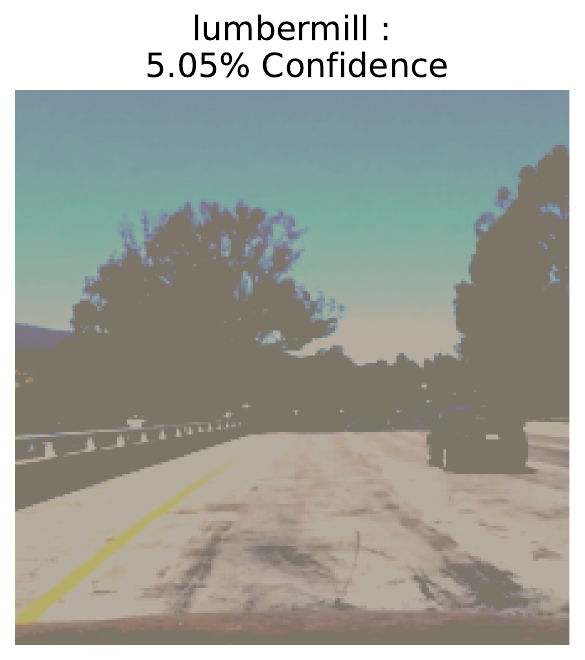}
    }
    \caption{Illustration of the prediction results of the ViT-B-32 model~\cite{pytorch2025vit}, trained on the ImageNet1K dataset, simulating autonomous driving scenarios. ViT is a Transformer-based model that uses self-attention. It splits the image into fixed-size patches and flattens them into tokens to capture global dependencies.}\label{fig:AAViT}
}
\end{figure*}

\begin{figure*}[!ht]
\vspace{-0.8cm}
\Revise{
    \captionsetup{format=hang,font=small, margin=2pt}
    \centering
    \subfloat[Input image is correctly classified as Ostrich.]{
        \includegraphics[width=0.23\linewidth, keepaspectratio]{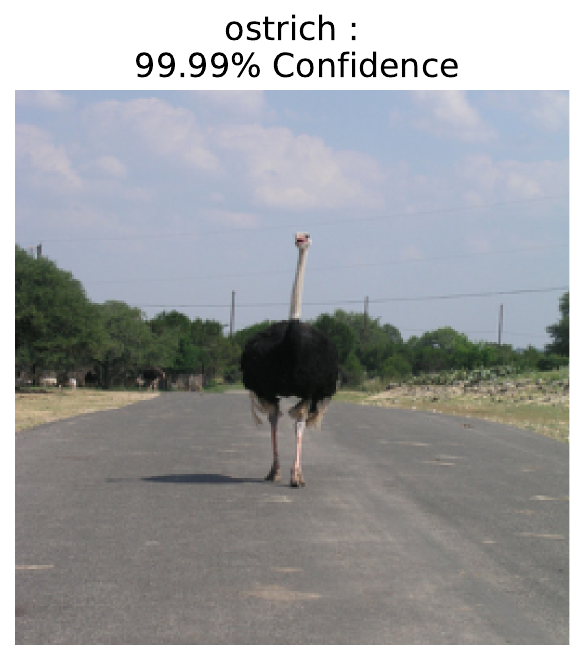}
    }
    \subfloat[Distributional shifts: rainy condition. Ostrich is misclassified as Geyser.]{
        \includegraphics[width=0.23\linewidth, keepaspectratio]{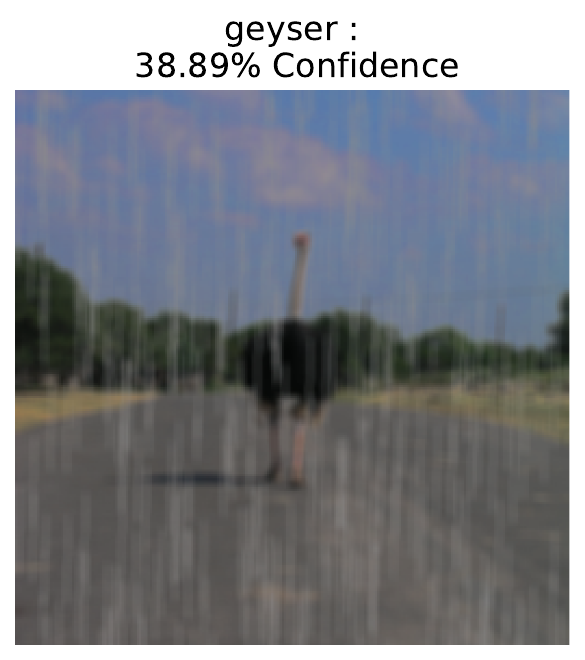}
    }
    \subfloat[Noisy and spatial translation: Lens glare and scaling. Ostrich is misclassified as Umbrella.]{
        \includegraphics[width=0.23\linewidth, keepaspectratio]{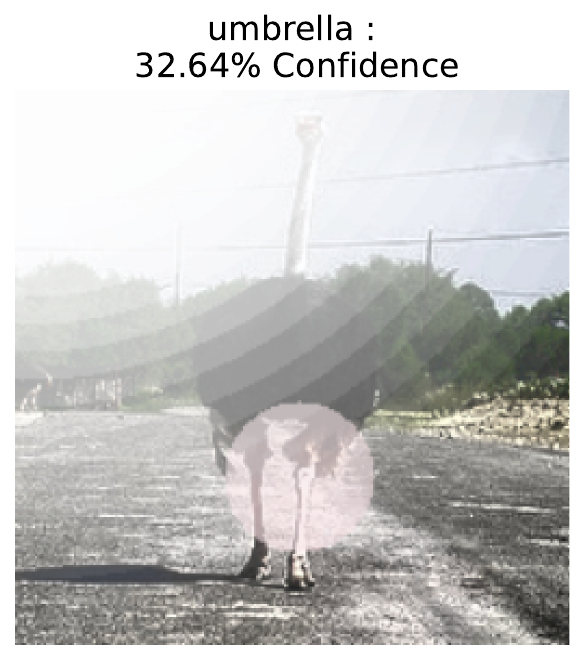}
    }
    \subfloat[Adversarial: PGD attack with $\epsilon=0.01$. Ostrich is misclassified as Maze.]{
        \includegraphics[width=0.23\linewidth, keepaspectratio]{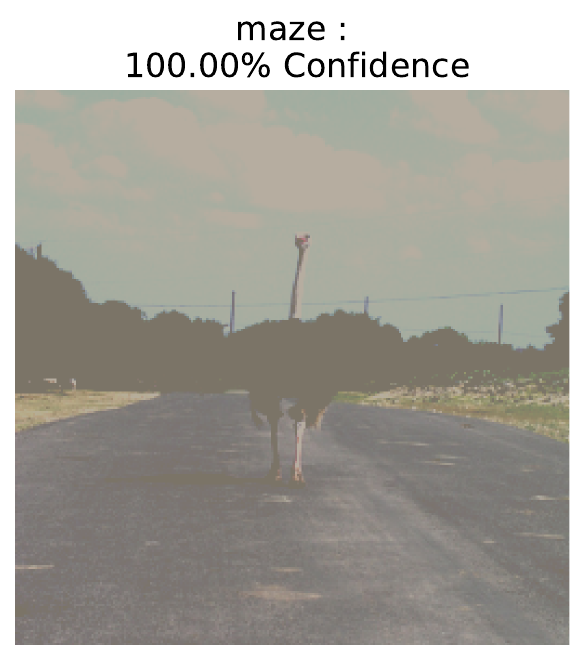}
    }
    \caption{Illustration of the prediction results of the ResNet-101~\cite{pytorch2025resnet}, trained on the ImageNet1K dataset, simulating real-world scenarios. ResNet-101 features residual connections (skip connections) that allow layers to learn residual functions (i.e., the difference between input and output), mitigating vanishing gradient issues.}\label{fig:AAResNet}
}
\end{figure*}

\begin{figure*}[!ht]
    \vspace{-0.8cm}
    \captionsetup{format=hang,font=small, margin=2pt}
    \centering
    \subfloat[Input image is correctly classified as Ostrich.]{
        \includegraphics[width=0.23\linewidth, keepaspectratio]{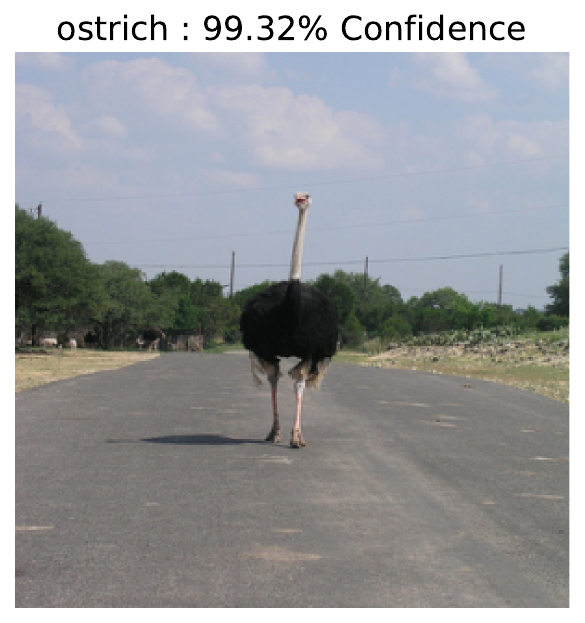}
    }
    \subfloat[Distributional shifts: rainy condition. Ostrich is misclassified as Lakeside.]{
        \includegraphics[width=0.23\linewidth, keepaspectratio]{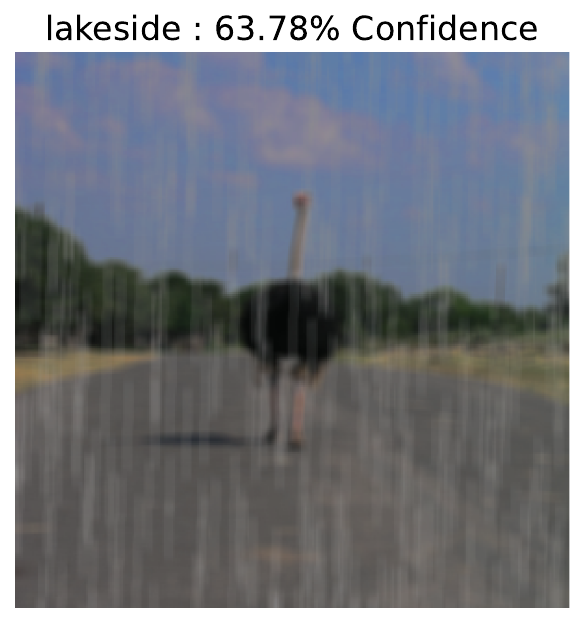}
    }
    \subfloat[Noisy and spatial translation: Lens glare and scaling. Ostrich is misclassified as Cock.]{
        \includegraphics[width=0.23\linewidth, keepaspectratio]{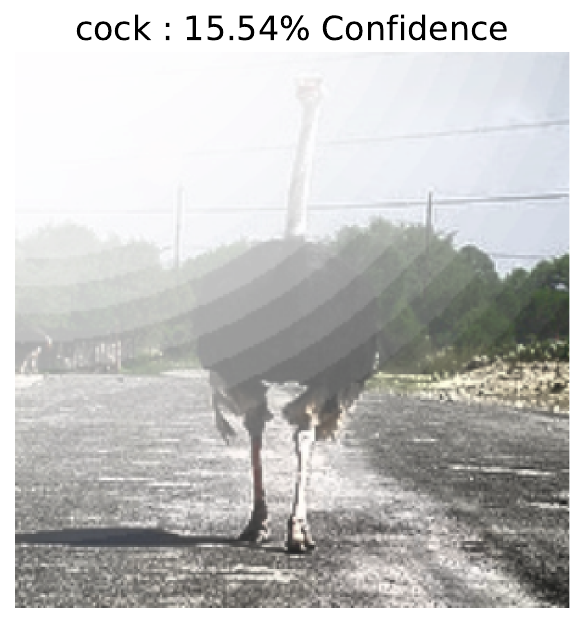}
    }
    \subfloat[Adversarial: PGD attack with $\epsilon=0.04$. Ostrich is misclassified as Sandbar.]{
        \includegraphics[width=0.23\linewidth, keepaspectratio]{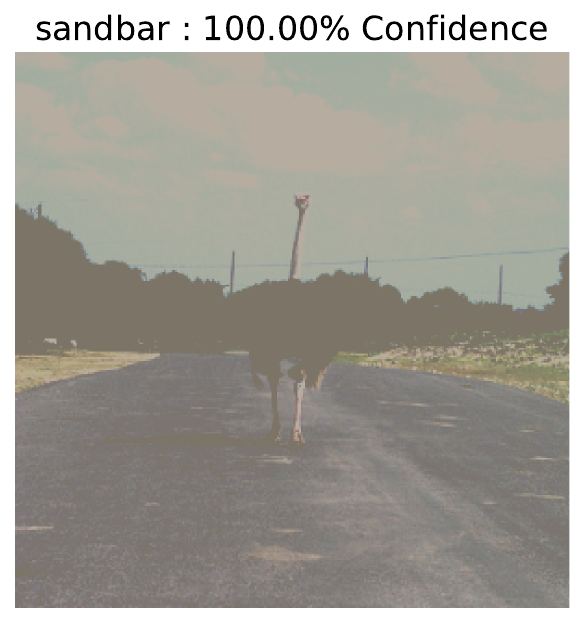}
    }
    \caption{Illustration of the prediction results of the Inception-V3 model~\cite{pytorch2024inception}, trained on the ImageNet1K dataset, simulating real-world scenarios. Inception-V3 employs label smoothing, factorized 7 x 7 convolutions, and an auxiliary classifier to propagate label information to lower layers of the network.}\label{fig:AAInception}
\end{figure*}

The expanded scope must adaptively ensure network robustness against \textit{distributional shifts} (refer to~\ref{sec:DSR}), \textit{noisy and spatially distorted inputs} (refer to~\ref{sec:NSR}), and \textit{adversarial attacks} (refer to~\ref{sec:AR}) leading to domain generalization. 
\Revise{
We simulate these three scenarios to demonstrate the robustness of the state-of-the-art ConvNeXT, ViT, ResNet, and Inception-V3 models, which are trained on ImageNet22K and ImageNet1K. The selection of ConvNeXT-XL models is primarily influenced by the 2025 DEEM release. DEEM, which uses ConvNeXT-B as its image encoder, claims to enhance the robustness, hallucination recognition, and foundational visual perception of LMMs. ViT, ResNet and Inception models serve as the backbone for many VLMs and VFMs listed in Table~\ref{tab:NetworksSummary}. ViT is a Transformer-based model that uses self-attention. ResNet introduces residual connections to enable training of deeper networks. Inception-V3 is a lightweight and computationally efficient model, suitable for mobile or edge devices. It can be observed in Figures~\ref{fig:AAConvNeXT},~\ref{fig:AAViT},~\ref{fig:AAResNet}, and~\ref{fig:AAInception} that the prediction results of the pre-trained ConvNeXT-XL~\cite{liu2022convnet}, ViT-B-32~\cite{pytorch2025vit}, ResNet-101~\cite{pytorch2025resnet} and Inception-V3~\cite{pytorch2024inception} models significantly differ from the original predictions when exposed to these three scenarios, highlighting the challenges of network robustness in real-world simulations. 
}

\subsection{Distributional shifts robustness}\label{sec:DSR}
Distributional shifts robustness is the ability of a network to maintain its performance when the distribution of the input data changes~\cite{shu2021encoding}. This includes variations in lighting, weather conditions, and image quality degradation. For example, a network trained on clean, sunny-day images may struggle with rainy conditions, highlighting the need for robust models capable of handling real-world scenarios~\cite{kang2024catch}. Distributional shifts can arise from semantic shifts (e.g., out-of-distribution (OOD) samples belonging to different classes) or covariate shifts (e.g., OOD samples originating from a distinct domain)~\cite{yang2024generalized}. Typically, studies~\cite{chen2024exploring} have characterized OOD detection as a binary classification problem. However, the assumption of a single ideal threshold that can perfectly separate in-distribution (ID) and OOD instances is often unrealistic in practice.

\subsection{Noisy and spatially distorted inputs robustness}\label{sec:NSR}
Noisy and spatially distorted inputs robustness is the ability of a network to maintain its performance when the input is corrupted by noise or errors and real-world appearance changes. Natural noise, arising from sources such as sensor damage and variation in environment, is an unavoidable and detrimental factor in computer vision~\cite{liu2024comprehensive}. Spatial distortions in inputs, arising from variations in the appearance of real-world objects, can be attributed to changes in viewpoint, non-rigid deformations of the objects, and transformations in perspective and scale within the scene~\cite{feng2023robust}.
Sensor damage can cause various image corruptions, including pixelation, blur, and noise~\cite{han2023interpreting}. Similarly, spatial transformations (e.g., rotation, scaling, shearing, and translation) can degrade network accuracy, including confidence levels.

\subsection{Adversarial robustness}\label{sec:AR}
Adversarial robustness refers to the network ability to maintain its performance when confronted with intentionally modified inputs, i.e., adversarial examples, designed to cause misclassification~\cite{meng2022adversarial}. This property is particularly critical for security-sensitive applications such as autonomous driving, border security, and medical diagnostics, where adversaries may deliberately manipulate inputs to compromise system reliability. Adversarial examples are digitally generated using various techniques~\cite{bitton2023evaluating, meyers2023safety}, depending on the adversary's knowledge of the target model, categorized as either white-box or black-box.

In white-box attacks, \textit{gradient-based methods}, such as the Fast Gradient Sign Method (FGSM)~\cite{goodfellow2015explaining}, and Projected Gradient Descent (PGD)~\cite{madry2018towards}, directly manipulate input data using model loss gradients. PGD is a multi-step variant of the FGSM algorithm that can generate more effective perturbations~\cite{ren2020adversarial}. \textit{Optimization-based techniques}, such as the Carlini and Wagner (C\&W) attack~\cite{carlini2017towards} and Broyden-Fletcher-Goldfarb-Shanno (L-BFGS) algorithm~\cite{szegedy2014intriguing}, involve finding an input that maximizes the model loss function. C\&W significantly improves upon L-BFGS by employing a more sophisticated loss functions, optimizing for a larger margin between the target class logit and the highest other logit. \textit{Gradient-approximation methods}, such as Backward Pass Differentiable Approximation (BPDA)~\cite{athalye2018obfuscated}, and Few-Pixel~\cite{shapira2023deep}, estimate gradients using techniques like finite differences, sampling methods, or other approximation algorithms to induce misclassification. BPDA can effectively generate adversarial examples even when defenses obscure gradients through masking or other techniques. Few-Pixel attacks are $\ell_0$ attacks that perturb only a very small number of input entries~\cite{shapira2023deep} in contrast to $\ell_\infty$ adversarial attacks, which perturb all input entries within a small bound ($\epsilon$). 

Black-box attacks generate adversarial examples by strategically manipulating inputs to exploit a target model output scores and decisions, often leveraging the transferability of these adversarial examples to other models~\cite{zhao2023remix}. Score-based (soft-label) attacks like zeroth-order gradient approximation is a derivative-free method used in black-box attacks to estimate the target model gradient using only its outputs~\cite{zhu2023zeroth}. When models only provide hard-label outputs (i.e, output probabilities or logits are absent), attacks can exploit decision boundaries as an alternative strategy to scores. Decision-based black-box attacks can be transferred from the digital domain to the physical world as they require only knowledge of the input layer and the output decision~\cite{jia2024fooling}. 

Pixel space and feature space can be directly exploited to perturb the pixels in the input image and the feature information of the image, respectively~\cite{zhu2024review}. In substitute model-based attacks, an adversary either gains access to the training data of the victim model or uses synthetic data to query the model for labeling~\cite{tao2023hard}. Subsequently, a substitute model is trained using the labeled data to approximate the victim model. Further, depending on the type of perturbation applied, both universal and specific attacks can be categorized as black-box attacks. Unlike sample-specific attacks that generate unique perturbations for each input, universal black-box adversarial attacks leverage transferable universal adversarial perturbations to generalize across diverse inputs. The universal adversarial patch attacks like HARDBEAT~\cite{tao2023hard} poses significant security concerns, as the adversary can generate a patch trigger for pre-trained models using gradient information.


\section{Network defense mechanisms}\label{sec:Defense}
\begin{figure*}[!ht]
    \centering
    \includegraphics[width=1\linewidth, keepaspectratio]{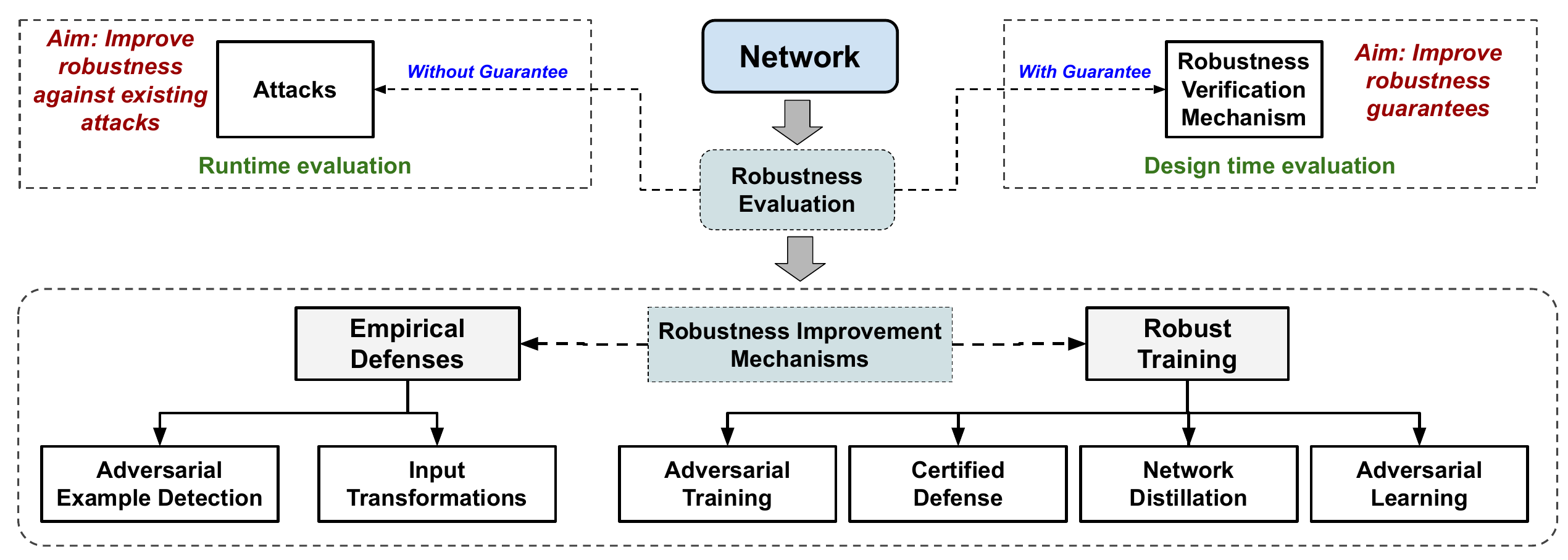}
    \caption{Illustration of networks robustness evaluation and improvement mechanisms}\label{fig:NetworkRobustness}
\end{figure*}

In response to \textbf{RQ2} (\textit{What are the prevalent mechanisms employed to improve network robustness}?), this section explores potential defense mechanisms employed to enhance network robustness. Figure~\ref{fig:NetworkRobustness} illustrate the networks robustness evaluation and improvement mechanisms. Network robustness can be improved using approaches such as empirical defenses and robust training~\cite{li2023sok}. Empirical defenses aim to enhance robustness against known attacks, while robust training approaches focus on strengthening robustness guarantees. Empirical defenses, such as \textit{adversarial detection}, and \textit{input transformations} are reactive approaches designed to detect attacks on networks after they have been constructed~\cite{yuan2019adversarial}. In contrast, robust training methods like \textit{adversarial training}, \textit{certified defense}, \textit{network distillation}, and \textit{adversarial learning}, are proactive approaches aimed at building networks that are resistant to attacks.

\subsection{Empirical defenses}
Adversarial example detection can be considered a passive defense strategy. Chen et al.~\cite{chen2022adversarial} describe that detection methods exploit unique properties of adversarial examples or utilize trained detectors to analyze inputs. Li et al.~\cite{li2024model} propose a model-agnostic detection method using high-frequency signals from adversarial noise. Pedraza et al.~\cite{pedraza2024leveraging} discuss the application of the chaos theory knowledge as an analogy to the perturbations caused by adversarial examples. In this context, adversarial perturbations are interpreted as points of chaos within various parts of the network, e.g., the input image or the inner features of the network layers. Consequently, properties and metrics from chaos theory, such as Lyapunov exponents (LEs), can be utilized as tools to detect potential adversarial examples.

Wang et al.~\cite{wang2021smsnet} propose the Stochastic Multifilter Statistical Network (SmsNet) for detecting adversarial examples. The original global pooling layer is replaced by a feature statistical layer that down-samples the input while simultaneously extracting statistical features. Gupta et al.~\cite{gupta2024visual} employ visual prompt engineering for detecting synthetic images to secure downstream tasks in face recognition system. Luo et al.~\cite{luo2022detecting} propose a Positive-Negative Detector (PNDetector) incorporating a classifier and a set of decision strategies to detect adversarial examples. PNDetector is an attack-agnostic method that detects adversarial examples by comparing the output similarity of their positive and negative representations.

Input transformations, such as data compression, filtering, or randomization, modify input data at runtime to mitigate adversarial examples~\cite{nesti2021detecting}. Xiong et al.~\cite{xiong2022towards} describe that JPEG compression can mitigate gradient-based adversarial attacks on image classifiers by removing high-frequency signal components within square image blocks. Sun et al.~\cite{sun2019adversarial} introduce sparse transformer layer to transform images such that corresponding clean and adversarial images have similar representations in both the quasi-natural image space and the learned feature space. The authors employ a convolutional dictionary learning-based method to construct the quasi-natural image space in an unsupervised manner. Projecting images onto this quasi-natural space achieves an optimal balance between preserving image details and effectively removing adversarial perturbations.

Xie et al.~\cite{xie2019feature} investigate feature denoising approaches to improve the robustness of network against adversarial attacks. They design convolutional network architectures that are equipped with building blocks designed to denoise feature maps. The building blocks use mean, median, and bilateral filters to enhance adversarial robustness. The study demonstrates that feature denoising based on non-local means achieves the best performance, aligning closely with principles of self-attention and non-local networks. Hu et al.~\cite{hu2024efficient} propose a CNN with heterogeneous kernels to extract multi-scale features for robust image denoising. The approach can provide a comprehensive representation of structural and contextual details, which is crucial for detail recovery and noise suppression. Additionally, an attention mechanism adaptively focuses on specific regions of the input image to adjust the weight of salient features, thereby enhancing the denoising effectiveness and efficiency.

Data randomization techniques mitigate adversarial attacks by applying random transformations to the input data~\cite{tiwari2022regroup}. Data randomization involving transformations like translation, rotation, clipping, scaling, and filling to the input sample, can be effective in defending against weak adversarial attacks but is less effective against more sophisticated attacks~\cite{wang2020deep}. Ma et al.~\cite{ma2024adversarial} investigate the impact of trainable and constrained random weights on adversarial robustness and propose incorporating random weights into the optimization process to fully exploit the potential of randomized defenses. The derivation of lower and upper bounds for random weight parameters is based on prediction bias and gradient similarity. The authors explain that randomness can increase gradient variability, but can also degrade overall performance.

\subsection{Robust training}
Adversarial training can be a highly effective defense against adversarial attacks that augments training data with adversarial examples~\cite{qian2022survey}. Several adversarial training variants such as bilateral, ensemble, feature scattering, and manifold are widely studied for designing robust networks. The fundamental concept of adversarial training is formulated as a minimax optimization problem, balancing inner maximization of adversarial perturbations and outer minimization of model loss~\cite{madry2018towards}. The inner phase maximizes loss to transform benign inputs into adversarial examples, and the outer phase focuses on minimizing loss until convergence. Rice et al.~\cite{rice2020overfitting} investigate techniques to prevent robust overfitting, a dominant phenomenon in adversarial robust training. They examine the effects of both implicit and explicit regularization methods on mitigating overfitting. Liu et al.~\cite{liu2024comprehensive} report that, although adversarial training improves adversarial robustness, it can degrade natural robustness for certain OOD datasets.

Shu et al.~\cite{shu2021encoding} propose adversarial training to enhance the robustness of CNNs against variations in image style and appearance, addressing challenges that extend beyond pixel-level perturbations. They introduce Adversarial Batch Normalization (AdvBN), a technique for network training that applies an adversarial feature shift by re-normalizing with the most adverse mean and variance values prior to each gradient update. The adversarial directions in feature space are iteratively computed using PGD on batch statistics, resulting in a trained network that is robust to domain shifts characterized by alterations in feature statistics. Jia et al.~\cite{jia2022adversarial} propose learnable attack strategy (LAS-AT) that learns to automatically produce  sample-dependent attack strategies to improve the model robustness. Wei et al.~\cite{wei2023cfa} propose Class-wise Calibrated Fair Adversarial Training (CFA) that dynamically adjusts adversarial configurations per class during training and modifies the weight averaging to enhance and stabilize the robustness of the least robust class.

Certified defense extends adversarial training to achieve robustness by propagating an input interval for a given perturbation radius through the network~\cite{frosio2023best}. The corresponding set of output bounds enables estimation of worst-case scenarios and derivation of the training cost function (e.g., worst-case entropy). Zhang~\cite{zhang2020machine} explains that certified defense techniques aim to minimize a robust loss, defined as the maximum loss within a specified perturbation set of each data point, to achieve a robust optimization. Certified defense techniques can be employed to train networks that are deterministically certified robust to $\ell_p$ norm bounded adversarial perturbations~\cite{xu2020automatic} and geometric image transformations~\cite{yang2023provable}.

Distillation is a training process that transfers knowledge from a larger, more complex network to a smaller, more efficient one, thereby reducing computational costs while maintaining performance. Papernot et al.~\cite{papernot2016distillation} propose using distillation as a defensive mechanism against adversarial examples. Dong et al.~\cite{dong2024robust} incorporate intermediate adversarial samples along the adversarial path for defensive distillation. An adaptive weighting mechanism calibrates the influence of each intermediate sample, facilitating the distillation of adversarial paths. This refinement determines a more robust decision boundary and minimizes an upper bound on the adversarially robust risk. Studies have reported that use of gradient masking or obfuscation strategies for defensive distillation can not fully mitigate attacks~\cite{chen2024learn}.

Liang et al.~\cite{liang2023advanced} propose a defense mechanism that enhances knowledge distillation by adding noise to the logit layer, preventing the adversary from directly recovering the genuine logits. Subsequently, they train an ensemble of networks, aggregating their outputs through vote ranking to mitigate potential performance degradation from noise injections. Kuang et al.~\cite{kuang2023improving} propose a robust soft-label distillation method to increase the mutual information between latent features and output predictions. They introduce an adaptive feature distillation method that automatically transfers relevant knowledge from teacher to student models, reducing input-latent feature mutual information.

Adversarial learning proactively integrates diverse attacks, including poisoning, evasion, and model theft, to build robust models~\cite{zhang2022towards}. Poisoning attacks introduce malicious data into training data to disrupt model training or retraining. Evasion attacks aim to mislead the model during inference by corrupting the input. Model theft involves replication of the model without direct access to its parameters or data. Qian et al.~\cite{qian2022survey} describe the use of adversarial techniques, such as generative adversarial networks, in adversarial learning to train models. The authors compare adversarial learning with adversarial training, stating that adversarial learning focuses more on improving model performance on standard data, while adversarial training prioritizes model robustness against adversarial attacks.

Fang et al.~\cite{fang2022enhanced} describe Generative Adversarial Networks (GANs) as an example of adversarial learning that consists of two competing models, namely, a generator and a discriminator. The generator produces samples that have visibly near identical characteristics to real data maximizing the discriminator. When the model converges, the discriminator loss function approaches a state of maximum entropy indicating the discriminator can not further distinguish between the real and generated samples. Leite and Xiao~\cite{leite2020improving} propose an adversarial learning approach to enhance cross-subject performance. The approach uses data augmentation to generate training data that mimics artificial subjects, thereby forcing the classifier to ignore subject-dependent information.

\section{Discussions and recommendations}\label{sec:Discussions}
This section elaborates \textbf{RQ3} (\textit{What challenges are associated with defense mechanisms employed to enhance network robustness}?), \textbf{RQ4} (\textit{What aspects, i.e., the properties and components, of a network should be analyzed for strengthening network robustness}?), and \textbf{RQ5} (\textit{What quantitative metrics can be used to benchmark network robustness}?) to advance the state-of-the-art in robust network training and empirical defense methods. Researchers and practitioners can utilize these findings to steer future research and development efforts. 

\subsection{RQ3: Challenges}
Robustness emphasizes proactive resilience to anticipated variations, though it is often conflated with safety and security~\cite{gupta2022non}. As illustrated in Figure~\ref{fig:AAConvNeXT},~\ref{fig:AAViT},~\ref{fig:AAResNet}, and~\ref{fig:AAInception}, \textbf{RQ1} posits the requirement for networks robustness against a range of input variations, including distributional shifts, noisy or spatially distorted data, and adversarial perturbations, to ensure reliable and widespread real-world deployment. \Revise{CLIP, DETR, and Mask2Former utilize ResNet as their image encoder, while ImageBind, DINOv2, and SAM adopt ViT for this purpose. DEEM and LLaVA employ ConvNeXT and CLIP, respectively, as image encoders. As discussed in Section~\ref{sec:VFMs}, the robustness of image encoders is essential for achieving reliable visual understanding and for minimizing visual hallucinations in VLMs.} Notably, black-box attacks, including transfer-based and query-based approaches, can severely compromise the robustness of VLMs by targeting the underlying encoders~\cite{zhao2023evaluating}. Transfer-based attacks employ surrogate models, under the white-box control of the adversary, to generate adversarial examples that are subsequently applied to victim models. In contrast, Query-based attacks repeatedly query victim models by providing image inputs and obtaining text outputs in order to estimate gradients or execute natural evolution algorithms.

\begin{figure*}[!ht]
    \centering
    \captionsetup{format=hang,font=small, margin=5pt}
    \hyphenpenalty 10000
    \subfloat[Performance against the FGSM attack. Accuracy degrades from 100\% to 89.4\%.]{
        \includegraphics[width=0.48\linewidth, keepaspectratio]{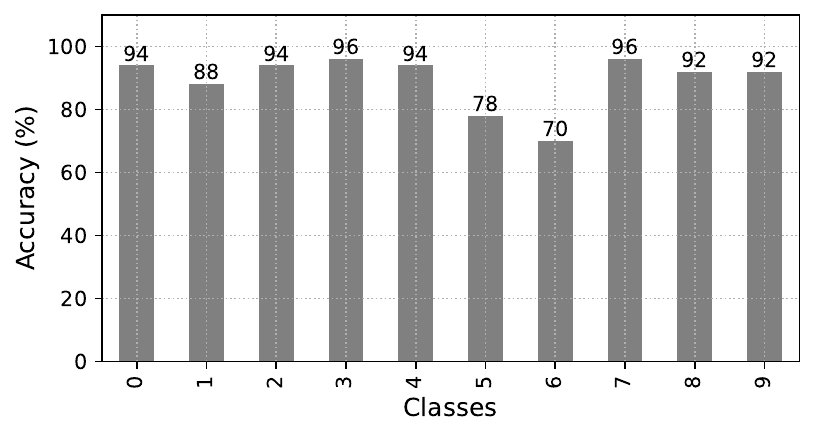}
    }
    \subfloat[Performance against the PGD attack. Accuracy degrades to 8.4\% from 100 \%]{
        \includegraphics[width=0.48\linewidth, keepaspectratio]{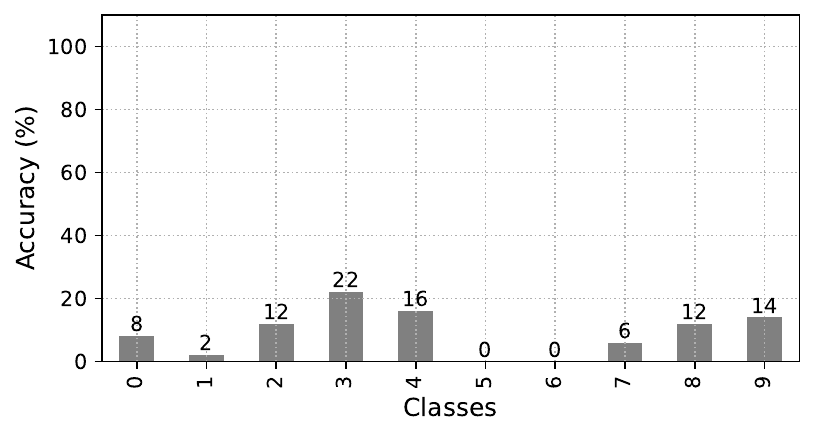}
    }
    \caption{ConvNeXT-L model (Total params: 197,740,264) performance against FGSM and PGD attacks.}\label{fig:ConvNeXTAdv}
\end{figure*}

Building on DEEM~\cite{luo2025deem}, a VLM released in 2025 that employs ConvNeXt-B for image encoding, we investigate the robustness of a larger, pre-trained ConvNeXt-L model\footnote{https://github.com/facebookresearch/ConvNeXt}. Similarly, to assess the robustness of multimodal encoders, we evaluate the more recent ImageBind encoder\footnote{\url{https://github.com/facebookresearch/ImageBind}} (released in 2023) in comparison to the CLIP image encoder (released in 2021) used in LLaVA~\cite{liu2023visual}. The performance of the ConvNeXT-L model is evaluated on a subset of the ImageNet dataset with 10 classes, where it achieves 100\% accuracy on the clean images. However, when the model is subjected to adversarial attacks with a small perturbation, its performance deteriorates. Figure~\ref{fig:ConvNeXTAdv} compares the performance of the ConvNeXT-L model, trained on the ImageNet1K dataset, against FGSM~\cite{goodfellow2015explaining} and PGD~\cite{madry2018towards} attacks with a perturbation radius of $0.1$ using $\ell_{\infty}$. 

\begin{figure}[!ht]
    \centering
    \captionsetup{format=hang,font=small, margin=5pt}
    \hyphenpenalty 10000
    \subfloat[10-classes embeddings with no attack\label{fig:DSSet4}]{
        \includegraphics[width=0.48\linewidth, keepaspectratio]{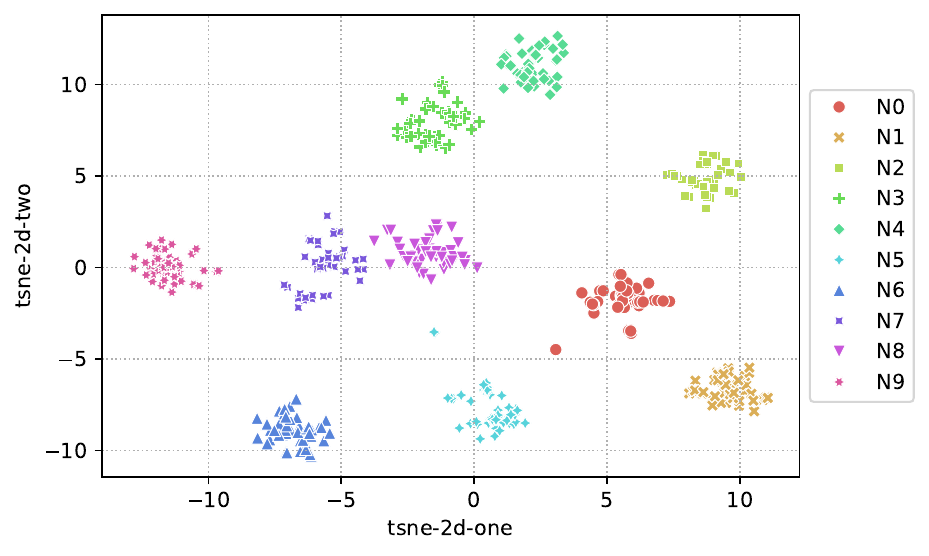}
    }
    \subfloat[10-classes embeddings after PGD attack\label{fig:DSSet6}]{
        \includegraphics[width=0.48\linewidth, keepaspectratio]{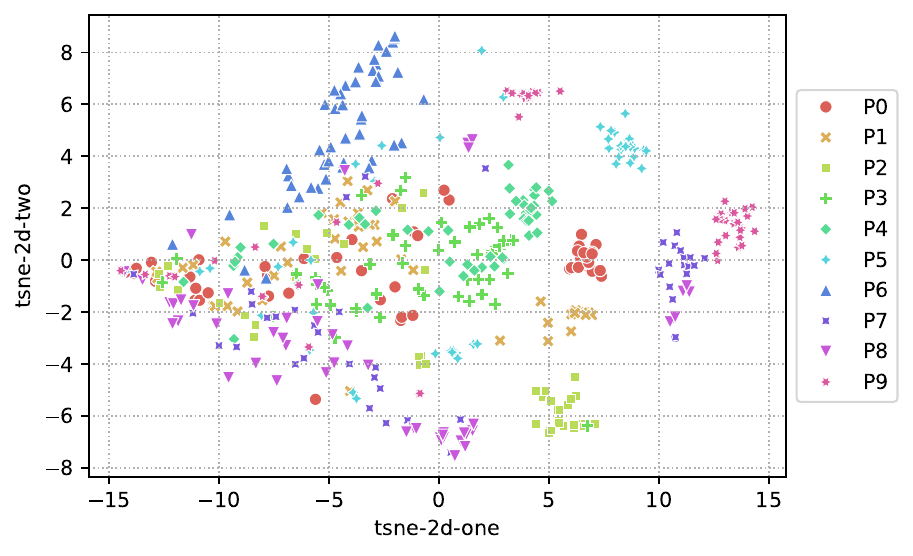}
    }
    \caption{t-SNE plots comparing the embeddings generated by the ImageBind model for a 10-class ImageNet subset (50 images per class) under no-attack condition and PGD attack.}\label{fig:ImageBindEmbComparison}
\end{figure}

ImageBind provides a unified joint embedding space $E_{\mathcal{I},\mathcal{M}}$, where $\mathcal{I}$ represents images and $\mathcal{M}$ denotes other modalities, such as text, audio, thermal, depth, and inertial measurement units (IMUs)~\cite{girdhar2023imagebind}. To generate adversarial examples under black-box conditions to evaluate ImageBind robustness, which uses ViT for image encoding, we employ ConvNeXT as it effectively bridges CNNs and Transformers. Figure~\ref{fig:ImageBindEmbComparison} compares the ImageBind embeddings of a 10-class subset of the ImageNet dataset under both clean and adversarial scenarios. The embeddings of the 10 classes are clearly distinguishable in the clean setting. However, when the same dataset is perturbed using a PGD attack with $\epsilon = 0.1$ under the $\ell_{\infty}$ norm, the embedding space collapses, making the classes indistinguishable. To demonstrate that, we design a multi-class classification model using SVM (Support Vector Machine), which shows a reduction in accuracy from 100\% on clean embeddings to 74.8\% on perturbed embeddings, highlighting the vulnerability of encoder performance to black-box attacks. VLMs rely heavily on aligned features across modalities for effective decision-making. Thus, robust encoders are crucial, as greater modality discrepancies can impair the quality of the learned representations.

\begin{table*}[!ht]
    \centering
    \footnotesize
    \hyphenpenalty 10000
    \caption{A comparison of defense strategies used for enhancing network robustness}\label{tab:DefenseTaxonomy}
    \begin{tblr}{
        width=1\linewidth,
        colspec = {p{.12\linewidth}  p{.22\linewidth} p{.4\linewidth}  p{.01\linewidth} p{.01\linewidth} p{.02\linewidth}},
    }\hline
    \SetCell[r=2]{c}{\textbf{Defense mechanism strategy}} & \SetCell[r=2]{c}{\textbf{Reference/Year}} & \SetCell[r=2]{c}{\textbf{Evaluation criteria}} & \SetCell[c=3]{c}{\textit{Use/Modify}}  \\\cline{4-6}
     &  &  & \rotatebox[origin=c]{90}{\textbf{Input}} & \rotatebox[origin=c]{90}{\textbf{Network}} & \rotatebox[origin=c]{90}{\parbox[l]{1cm}{\textbf{Training data}}} \\\hline
    \SetCell[r=3]{l}{Adversarial example detection} & SmsNet~\cite{wang2021smsnet}, 2021 & Adversarial images generated using FGSM, C\&W, DeepFool~\cite{moosavi2016deepfool}, and Basic Iterative Method (BIM)~\cite{kurakin2018adversarial}. & - & \Checkmark & -\\\hline
    & PNDetector~\cite{luo2022detecting}, 2022 & FGSM, L-BFGS, C\&W, DeepFool, Elastic-Net~\cite{chen2018ead}, SPSA~\cite{uesato2018adversarial}, Jacobian-based Saliency Map Attack (JSMA)~\cite{papernot2016limitations} & - & \Checkmark & - \\\hline
    & Visual prompt engineering~\cite{gupta2024visual}, 2024 & Synthetic images generated using GAN. & - & \Checkmark & -\\\hline
    \SetCell[r=5]{l}{Input transformations} & Data compression methods~\cite{xiong2022towards}, 2022 & Gradient-based adversarial attacks like FGSM & \Checkmark & - & -\\\hline
    & Sparse Transformation Layer~\cite{sun2019adversarial}, 2019 & Adversarial images generated using FGSM, BIM, DeepFool, C\&W & \Checkmark & - & -\\\hline
    & Image denoising~\cite{xie2019feature}, 2019; ~\cite{hu2024efficient}, 2024 & PGD, CAAD black-box setting, real-world noisy images & \Checkmark & - & -\\\hline
    & Data randomization~\cite{tiwari2022regroup}, 2022; ~\cite{wang2020deep}, 2020 & PGD, DeepFool, BIM, C\&W, SPSA, Boundary, and Spatial & - & - & \Checkmark\\\hline
    & Random weights~\cite{ma2024adversarial}, 2024 & PGD & - & \Checkmark & - \\\hline
    \SetCell[r=4]{l}{Adversarial training} & Minimax optimization~\cite{madry2018towards}, 2018 & PGD, FGSM, C\&W & - & \Checkmark & \Checkmark \\\hline
    & AdvBN~\cite{shu2021encoding}, 2021 & Distributional shifts using ImageNet-C, ImageNet-Ins, ImageNet-Sketch, ImageNet-Style & - & \Checkmark & - \\\hline
    & LAS-AT~\cite{jia2022adversarial}, 2022 & PGD, C\&W, AutoAttack & - & \Checkmark & \Checkmark \\\hline
    & CFA~\cite{wei2023cfa}, 2023 & AutoAttack & - & - & \Checkmark\\\hline
    \SetCell[r=3]{l}{Certified defense} & CROWN~\cite{zhang2018efficient}, 2018 & PGD & - & \Checkmark & \Checkmark \\\hline
    & IBP~\cite{gowal2019scalable}, 2019 & PGD & - & \Checkmark & \Checkmark \\\hline
    & AutoLirpa~\cite{xu2020automatic}, 2020 & PGD & - & \Checkmark & \Checkmark \\\hline 
    \SetCell[r=4]{l}{Network distillation} & Defensive distillation~\cite{papernot2016distillation}, 2016 & JSMA & - & \Checkmark & -\\\hline
    & Ensemble network with noisy logit~\cite{liang2023advanced}, 2023 & L-BFGS, C\&W & - & \Checkmark & -\\\hline
     &  Information Bottleneck Distillation~\cite{kuang2023improving}, 2023 & PGD, AutoAttack & - & \Checkmark & -\\\hline
     & Knowledge distillation~\cite{dong2024robust}, 2024 & Natural, PGD, AutoAttack & - & \Checkmark & -\\\hline
    \SetCell[r=2]{l}{Adversarial learning} & Cross-subject performance~\cite{leite2020improving}, 2020 & HARD, HARD2, PAMAP2~\cite{reiss2012introducing} & - & \Checkmark & -\\\hline
     & AdvFocusDMTL~\cite{fang2022enhanced}, 2022 & CelebA~\cite{liu2015deep}, DukeMTMC-attribute~\cite{lin2019improving}, RAF-DB~\cite{li2017reliable} & - & \Checkmark & -\\\hline
    \end{tblr}
\end{table*}

The comparative analysis presented in Table~\ref{tab:DefenseTaxonomy} demonstrates that current defense mechanisms predominantly focus on a limited range of attacks to evaluate robustness, thereby revealing their shortcomings in comprehensively addressing \textbf{RQ1}. Particularly, they insufficiently account for natural variations. Current research on adversarial example detection methods, adversarial training strategies, and certified defense mechanisms primarily investigate a network robustness against $\ell_p$ norm-bounded adversarial attacks, such as FGSM, PGD, and C\&W. However, attacks like semantic \textit{similarity attacks} based on the similarity of feature representations~\cite{luo2022frequency}, \textit{spatial transformation attacks} that minimize local geometric distortions rather than the $\ell_p$ pixel error between adversarial and original instances~\cite{xiao2018spatially}, and \textit{composite adversarial attacks} where an attack policy is formed by serially connecting multiple attackers, with each attack output serving as the initialization input for the next~\cite{mao2021composite} demand deeper investigation.

Evasion attack detection methods, such as PNDetector~\cite{luo2022detecting} and Visual Prompt Engineering~\cite{gupta2024visual}, can be incorporated externally into the pipeline to perform sanity checks on input data, can enhance the robustness of VFMs for downstream tasks without affecting their inference performance. Similarly, input transformation mechanisms can detect adversarial examples at runtime by modifying input data, providing a simple and computationally effective way to mitigate adversarial attacks without changing or retraining the neural network~\cite{nesti2021detecting}. However, mechanisms such as SmsNet~\cite{wang2021smsnet}, which relies on the statistical properties of samples for adversarial example detection, and Minimax optimization~\cite{madry2018towards}, which augments the training data with adversarial examples generated using PGD, can introduce additional computational overhead during VFM training. The robustness-accuracy trade-off in adversarial training can degrade the zero-shot capabilities of VLMs by inadvertently introducing a substantial domain shift to boost robustness~\cite{luo2024enhancing}.

Certified defense mechanisms such as CROWN~\cite{zhang2018efficient}, IBP~\cite{gowal2019scalable}, and AutoLirpa~\cite{xu2020automatic} formally verify a network robustness guarantees against a predefined class of adversarial attacks, however, this often comes at the cost of reduced network accuracy. Knowledge distillation can be an effective technique for obtaining lightweight models with reduced computational resource demands and enhanced inference speed, without significantly compromising performance~\cite{chen2024learn}. Techniques such as adversarial learning and training can be integrated into a multi-teacher knowledge distillation approach to balance the performance and robustness of lightweight models. \Revise{Our investigation provides a roadmap for enhancing network robustness by identifying vulnerabilities and evaluating defense mechanisms.} To effectively address \textbf{RQ3}, future research can prioritize datasets that incorporate distributional shifts, noisy or spatially distorted data, and other realistic corruptions and physical attacks for a more accurate evaluation of network robustness.

\subsection{RQ4: Ablation approach}
Ablation~\cite{shu2021encoding, kuang2023improving} can be highly useful for assessing the contributions of different training techniques, tricks, and model components. The approach involves initially training a baseline model with all techniques or tricks applied. Subsequently, each technique is systematically removed or disabled, and the model is retrained to assess the significance and contribution of each technique to the overall robustness. For example, consider training a model using adversarial techniques that incorporate PGD-generated adversarial examples, combined with weight decay for regularization and random cropping for data augmentation. The steps for the ablation study are outlined below. The contribution of each technique to improving robustness can be quantified by comparing the results of the ablation experiments.
\begin{itemize}[leftmargin=*,itemsep=3pt]
\item \textit{Full training}: Training with PGD attacks, random cropping, and weight decay.
\item \textit{Ablation of weight decay}: Training with PGD attacks and random cropping, but without weight decay.
\item \textit{Ablation of random cropping}: Training with PGD attacks and weight decay, but without random cropping.
\item \textit{Ablation of PGD attacks}: Standard training with weight decay and random cropping (i.e., no adversarial training).
\end{itemize}

\begin{figure}[!ht]
    \centering
    \includegraphics[width=.8\linewidth, keepaspectratio]{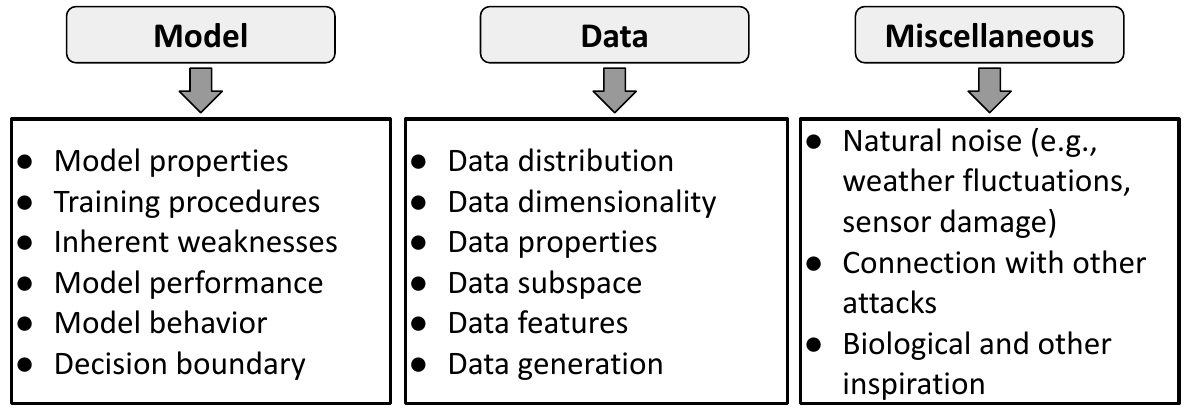}
    \caption{Illustration of model-centric, data-centric, and miscellaneous perspectives to derive a structured framework for ablation studies~\cite{gupta2024visual}.}\label{fig:NetworkBuildingBlocks}
\end{figure}

As illustrated in Figure~\ref{fig:NetworkBuildingBlocks}, we examine model-centric, data-centric, and miscellaneous perspectives that can be used to derive a structured framework for ablation studies. 
Understanding each perspective can comprehend model behavior, identify redundancies, and generate actionable benchmarks towards the robustness of defenses.

The \textit{model perspective} encompasses network characteristics exploitable for adversarial example generation, such as model properties, i.e., linearity hypothesis, model architecture, or activation function selection~\cite{carlini2017towards}. Training procedures like updating parameters using gradient descent, weight initialization, training loss calculation, regularization, and batch normalization can be some direct targets~\cite{noack2021empirical}. Prediction-, Epistemic- and Aleatoric uncertainty can be the viewpoints to interpret inherent weaknesses~\cite{hullermeier2021aleatoric}. Model performance must include clean-, corruption- and transformation accuracy~\cite{li2020learnable}. Similarly, model behavior must analyze internal modules, i.e., the activation of critical neurons, channel-wise activation, and statistical characteristics of hidden layers output~\cite{guo2023comprehensive}. Lastly, the decision boundary can encompass the boundary tilting hypothesis, the decision boundary similarity among models, the curvature of the decision boundary, and the decision surface to strengthen a model robustness~\cite{serban2020adversarial}.

The \textit{data perspective} can analyze aspects such as data distribution, dimensionality, properties, subspaces, features, and GAN-generated images to understand the presence or generation of adversarial examples~\cite{han2023interpreting}. Data distribution used for training and evaluating CNNs must be selected reasonably to avoid unexpected behavior for out-of-distribution data, distribution weaknesses, and differences~\cite{feng2023dynamic, gupta2025evaluating}. Further, data dimensionality, i.e., the high dimension of the data, can be a significant contributor to a model fragility owing to intricate and convoluted decision boundaries~\cite{li2020learnable}. Consequently, adversarial examples can easily exploit complex decision boundaries to identify small perturbations that can cause significant changes in the model predictions~\cite{kienitz2022comparing}. Data properties like the number of classes in the dataset and the sample complexity of the dataset can impact the generation and effectiveness of adversarial examples. In simple words, insufficient training data or limited generalization ability can make models more vulnerable to adversarial examples~\cite{chen2021local}.

As discussed previously, the high dimension of the input data is naturally a hindrance to exploring the existence of adversarial examples. Studies~\cite{li2020learnable, li2020adversarial} have shown that projecting the inputs into a new subspace or a low-dimension subspace can be useful in investigating adversarial examples. Thus, abnormal patterns in data subspace, adversarial subspace properties, and data manifold can explain CNNs vulnerability~\cite{jiang2022attacks}. Similarly, the identification of adversarial examples and clean images can be enhanced through the analysis of data characteristics. This includes the examination of abnormal patterns at the feature level using visualization methods like t-SNE, feature attribution maps, intra-class and inter-class feature classification, as well as the scrutiny of local and global features and high-level semantic features detected by neurons~\cite{fu2023multi}. Lastly, for data generation, GANs can produce synthetic samples that closely resemble the real data distribution~\cite{karras2020analyzing,gupta2025evaluating}. By observing the generation process and the differences between real and generated samples, one can gain insights into the factors that contribute to the generation of adversarial examples.

The \textit{miscellaneous perspective} includes natural noise arising from sources such as weather fluctuations and sensor damage~\cite{liu2024comprehensive}. It also includes connection with other types of attacks like poisoning and backdoor attacks~\cite{zhou2022adversarial} and inspiration from biological and other fields~\cite{han2023interpreting}, e.g., knowledge transfer, game theory, algebraic topology, information geometry, and model-based diagnosis and knowledge compilation.

\subsection{RQ5: Metrics}
Traditional neural network verification methods typically focus on prediction accuracy and validation metrics like efficiency, which alone are inadequate for effectively quantifying robustness~\cite{meng2022adversarial}. We address this gap by describing relevant benchmarking metrics for assessing network robustness listed in Table~\ref{tab:Metrics}. Network robustness can be measured using both static (i.e., design-time) and dynamic (i.e., runtime) metrics, leveraging data-centric and network-centric approaches~\cite{guo2023comprehensive}. Data-centric approaches can be useful for measuring the integrity of test examples by considering the neuron coverage and data visual imperceptibility. Network-centric evaluation metrics can be useful for measuring the model performance in the adversarial setting. 
\begin{table}[!ht]
    \centering
    \footnotesize
    \hyphenpenalty 10000
    \caption{List of metrics to benchmark network robustness using data and model centric approaches}\label{tab:Metrics}
    \begin{tabularx}{1\linewidth}{p{.26\linewidth} p{.64\linewidth}}\hline
     \multicolumn{2}{c}{\textbf{Data-centric}}\\\hline
     Neuron coverage & k-Multisection Neuron Coverage, Neuron boundary coverage \\\hline
     Data imperceptibility & Average $\ell_p$ Distortion, Average structural similarity, Perturbation sensitivity distance \\\hline
     \multicolumn{2}{c}{\textbf{Network-centric}}\\\hline
     Task Performance & Clean accuracy\\\hline
     Adversarial performance & Adversarial accuracy on white-box attacks, Adversarial accuracy on black-box attacks, Average confidence of adversarial class, Average confidence of true class, Noise tolerance estimation \\\hline
     Corruption Performance & Mean corruption error (mCE), Relative mCE \\\hline
     Defense Performance & Classification Accuracy Variance, Classification Confidence Variance\\\hline
     Boundary-based & Empirical boundary distance 1 and 2 \\\hline
     Consistency-based & Empirical noise insensitivity\\\hline
     Neuron-based & Neuron sensitivity and uncertainty\\\hline
    \end{tabularx}
\end{table}

A binary-valued function $R(f|x,y)$ can be used to determine whether a model is robust for an example-label pair ($x, y$)~\cite{zhang2022towards}. Specifically, $R(f|x,y) = 1$ when a network ($f$) is guaranteed to be robust on $x$, provided the outputs are consistent with the condition defined in Equation~\eqref{eq:Zhang}.
\begin{equation}\label{eq:Zhang}
    f(x') = f(x), s.t. \parallel x' - x \parallel_p
\end{equation}
Otherwise, the outputs are inconsistent, and $R(f|x,y) = 0$. Consequently, when aiming for $R(f|x,y) = 1$, the model must produce consistent outputs for different perturbations. Model robustness can be assessed by plotting model accuracy against different perturbation budgets~\cite{liu2024comprehensive}. The minimum perturbation budget for a given sample is determined using binary search, which, when applied to a crafted adversarial example under $\ell_p$ norm constraints, results in misclassification. Then, the percentage of data samples with minimum perturbations smaller than each $\epsilon$ is computed to plot the curve.

Equation~\eqref{eq:CorruptionError} presents the Corruption Error ($CE$) for measuring standardized aggregate performance~\cite{hendrycks2018benchmarking}. $E$ is the top-1 error rate, indicating the frequency of incorrect highest-probability predictions of the model. $f$ is a trained model for evaluation, $f_{baseline}$ is a baseline model, $c$ is the corruption type and $s$ is the severity level ($1 \leq s \leq 5$). The mean CE can be computed by averaging $CE$ for each corruption type.
\begin{equation}\label{eq:CorruptionError}
    CE^f_C = \frac{\Sigma^5_{s=1} E^f_{s,c}}{\Sigma^5_{s=1} E^{f_{baseline}}_{s,c}}
\end{equation}

Equation~\eqref{eq:FoolingRate} defines the Fooling Rate ($FR$) for evaluating model robustness~\cite{gu2022evaluating}. It is a ratio of correct predictions on clean images ($P$) to those on perturbed images ($Q$) under natural patch corruption or adversarial patch attacks.
\begin{equation}\label{eq:FoolingRate}
    FR = \frac{P}{Q}
\end{equation}

Similar to the Fooling Rate, Equation~\eqref{eq:RetentionRate} defines the retention rate ($Ret R$) to evaluate model robustness~\cite{zhou2022understanding}. It is a ratio of the model prediction correctness ($PR$) for the corrupted image dataset compared to the clean image dataset.
\begin{equation}\label{eq:RetentionRate}
    Ret R = \frac{PR_{corrupt}}{PR_{clean}}
\end{equation}

Formal verification using Interval Bound Propagation (IBP), which is based on interval arithmetic, assigns intervals that bound the minimum and maximum values of each input pixel and neuron in the network, resulting in computed lower and upper bounds for a given perturbed input ($x'$)~\cite{gowal2019scalable}. A network can robustly classify the input, if the lower bound ($\underline{f_y}$) of the true class ($y$) is greater than the upper bound ($\overline{f_j}$) of the rest of the classes ($j$) as shown in Equation~\eqref{eq:FVLPNorm}.
\begin{equation}\label{eq:FVLPNorm}
    \underline{f_y} (x') > \overline{f_j}(x') \quad \forall j \neq y
\end{equation}

Weng~\cite{weng2018towards} proposes CLEVER (Cross Lipschitz Extreme Value for Network Robustness), an estimation of the lower bound of the minimum perturbation that is independent of attack algorithms. Xu et al.~\cite{xu2020automatic} demonstrate that certified lower and upper bounds on network outputs, derived from input perturbations, can provide provable robustness guarantees.

\section{Conclusions}\label{sec:Conclusions}
Network robustness is crucial for the successful deployment of computer vision systems in real-world scenarios. While foundation models demonstrate high accuracy on benchmark datasets, their ability to maintain robust performance in the presence of data transformations, noise, erroneous inputs, and targeted adversarial attacks remains a significant and persistent challenge. Our investigation of empirical defense mechanisms, such as adversarial detection, and input transformations, and robust training mechanisms, such as adversarial training, certified defense, network distillation, and adversarial learning, employed to enhance network robustness, shows that they focus on a limited range of attacks when evaluating network robustness. It is recommended to thoroughly evaluate network robustness against requirements such as distributional shifts, noisy and spatially distorted inputs, and adversarial perturbations. Addressing these requirements is vital to ensuring reliable performance and broadening the applicability of computer vision to diverse and sensitive domains.

\bibstyle{sn-mathphys-num}
\bibliography{references}


\begin{thebibliography}{129}
\ifx \bisbn   \undefined \def \bisbn  #1{ISBN #1}\fi
\ifx \binits  \undefined \def \binits#1{#1}\fi
\ifx \bauthor  \undefined \def \bauthor#1{#1}\fi
\ifx \batitle  \undefined \def \batitle#1{#1}\fi
\ifx \bjtitle  \undefined \def \bjtitle#1{#1}\fi
\ifx \bvolume  \undefined \def \bvolume#1{\textbf{#1}}\fi
\ifx \byear  \undefined \def \byear#1{#1}\fi
\ifx \bissue  \undefined \def \bissue#1{#1}\fi
\ifx \bfpage  \undefined \def \bfpage#1{#1}\fi
\ifx \blpage  \undefined \def \blpage #1{#1}\fi
\ifx \burl  \undefined \def \burl#1{\textsf{#1}}\fi
\ifx \doiurl  \undefined \def \doiurl#1{\url{https://doi.org/#1}}\fi
\ifx \betal  \undefined \def \betal{\textit{et al.}}\fi
\ifx \binstitute  \undefined \def \binstitute#1{#1}\fi
\ifx \binstitutionaled  \undefined \def \binstitutionaled#1{#1}\fi
\ifx \bctitle  \undefined \def \bctitle#1{#1}\fi
\ifx \beditor  \undefined \def \beditor#1{#1}\fi
\ifx \bpublisher  \undefined \def \bpublisher#1{#1}\fi
\ifx \bbtitle  \undefined \def \bbtitle#1{#1}\fi
\ifx \bedition  \undefined \def \bedition#1{#1}\fi
\ifx \bseriesno  \undefined \def \bseriesno#1{#1}\fi
\ifx \blocation  \undefined \def \blocation#1{#1}\fi
\ifx \bsertitle  \undefined \def \bsertitle#1{#1}\fi
\ifx \bsnm \undefined \def \bsnm#1{#1}\fi
\ifx \bsuffix \undefined \def \bsuffix#1{#1}\fi
\ifx \bparticle \undefined \def \bparticle#1{#1}\fi
\ifx \barticle \undefined \def \barticle#1{#1}\fi
\bibcommenthead
\ifx \bconfdate \undefined \def \bconfdate #1{#1}\fi
\ifx \botherref \undefined \def \botherref #1{#1}\fi
\ifx \url \undefined \def \url#1{\textsf{#1}}\fi
\ifx \bchapter \undefined \def \bchapter#1{#1}\fi
\ifx \bbook \undefined \def \bbook#1{#1}\fi
\ifx \bcomment \undefined \def \bcomment#1{#1}\fi
\ifx \oauthor \undefined \def \oauthor#1{#1}\fi
\ifx \citeauthoryear \undefined \def \citeauthoryear#1{#1}\fi
\ifx \endbibitem  \undefined \def \endbibitem {}\fi
\ifx \bconflocation  \undefined \def \bconflocation#1{#1}\fi
\ifx \arxivurl  \undefined \def \arxivurl#1{\textsf{#1}}\fi
\csname PreBibitemsHook\endcsname

\bibitem[\protect\citeauthoryear{Gupta et~al.}{2024}]{gupta2024visual}
\begin{botherref}
\oauthor{\bsnm{Gupta}, \binits{S.}},
\oauthor{\bsnm{Raja}, \binits{K.}},
\oauthor{\bsnm{Passerone}, \binits{R.}}:
Visual prompt engineering for enhancing facial recognition systems robustness against evasion attacks.
IEEE Access
(2024)
\end{botherref}
\endbibitem

\bibitem[\protect\citeauthoryear{Liu et~al.}{2024}]{liu2024few}
\begin{barticle}
\bauthor{\bsnm{Liu}, \binits{F.}},
\bauthor{\bsnm{Zhang}, \binits{T.}},
\bauthor{\bsnm{Dai}, \binits{W.}},
\bauthor{\bsnm{Zhang}, \binits{C.}},
\bauthor{\bsnm{Cai}, \binits{W.}},
\bauthor{\bsnm{Zhou}, \binits{X.}},
\bauthor{\bsnm{Chen}, \binits{D.}}:
\batitle{Few-shot adaptation of multi-modal foundation models: A survey}.
\bjtitle{Artificial Intelligence Review}
\bvolume{57}(\bissue{10}),
\bfpage{268}
(\byear{2024})
\end{barticle}
\endbibitem

\bibitem[\protect\citeauthoryear{LeCun et~al.}{1998}]{lecun1998gradient}
\begin{barticle}
\bauthor{\bsnm{LeCun}, \binits{Y.}},
\bauthor{\bsnm{Bottou}, \binits{L.}},
\bauthor{\bsnm{Bengio}, \binits{Y.}},
\bauthor{\bsnm{Haffner}, \binits{P.}}:
\batitle{Gradient-based learning applied to document recognition}.
\bjtitle{Proceedings of the IEEE}
\bvolume{86}(\bissue{11}),
\bfpage{2278}--\blpage{2324}
(\byear{1998})
\end{barticle}
\endbibitem

\bibitem[\protect\citeauthoryear{Krizhevsky et~al.}{2012}]{krizhevsky2012imagenet}
\begin{botherref}
\oauthor{\bsnm{Krizhevsky}, \binits{A.}},
\oauthor{\bsnm{Sutskever}, \binits{I.}},
\oauthor{\bsnm{Hinton}, \binits{G.E.}}:
Imagenet classification with deep convolutional neural networks.
Advances in neural information processing systems
\textbf{25}
(2012)
\end{botherref}
\endbibitem

\bibitem[\protect\citeauthoryear{Simonyan and Zisserman}{2015}]{simonyan2015vggnet}
\begin{bchapter}
\bauthor{\bsnm{Simonyan}, \binits{K.}},
\bauthor{\bsnm{Zisserman}, \binits{A.}}:
\bctitle{Very deep convolutional networks for large-scale image recognition}.
In: \bbtitle{Proceedings of the 3rd International Conference on Learning Representations},
pp. \bfpage{1}--\blpage{14}
(\byear{2015})
\end{bchapter}
\endbibitem

\bibitem[\protect\citeauthoryear{Ren et~al.}{2016}]{ren2016deep}
\begin{bchapter}
\bauthor{\bsnm{Ren}, \binits{S.}},
\bauthor{\bsnm{Sun}, \binits{J.}},
\bauthor{\bsnm{He}, \binits{K.}},
\bauthor{\bsnm{Zhang}, \binits{X.}}:
\bctitle{Deep residual learning for image recognition}.
In: \bbtitle{Proceedings of the IEEE Conference on Computer Vision and Pattern Recognition},
vol. \bseriesno{3},
pp. \bfpage{770}--\blpage{778}
(\byear{2016})
\end{bchapter}
\endbibitem

\bibitem[\protect\citeauthoryear{Szegedy et~al.}{2016}]{szegedy2016rethinking}
\begin{bchapter}
\bauthor{\bsnm{Szegedy}, \binits{C.}},
\bauthor{\bsnm{Vanhoucke}, \binits{V.}},
\bauthor{\bsnm{Ioffe}, \binits{S.}},
\bauthor{\bsnm{Shlens}, \binits{J.}},
\bauthor{\bsnm{Wojna}, \binits{Z.}}:
\bctitle{Rethinking the inception architecture for computer vision}.
In: \bbtitle{Proceedings of the IEEE Conference on Computer Vision and Pattern Recognition},
pp. \bfpage{2818}--\blpage{2826}
(\byear{2016})
\end{bchapter}
\endbibitem

\bibitem[\protect\citeauthoryear{Huang et~al.}{2017}]{huang2017densely}
\begin{bchapter}
\bauthor{\bsnm{Huang}, \binits{G.}},
\bauthor{\bsnm{Liu}, \binits{Z.}},
\bauthor{\bsnm{Van Der~Maaten}, \binits{L.}},
\bauthor{\bsnm{Weinberger}, \binits{K.Q.}}:
\bctitle{Densely connected convolutional networks}.
In: \bbtitle{Proceedings of the IEEE Conference on Computer Vision and Pattern Recognition},
pp. \bfpage{4700}--\blpage{4708}
(\byear{2017})
\end{bchapter}
\endbibitem

\bibitem[\protect\citeauthoryear{Redmon}{2016}]{redmon2016you}
\begin{bchapter}
\bauthor{\bsnm{Redmon}, \binits{J.}}:
\bctitle{You only look once: Unified, real-time object detection}.
In: \bbtitle{Proceedings of the IEEE Conference on Computer Vision and Pattern Recognition},
pp. \bfpage{1}--\blpage{10}
(\byear{2016})
\end{bchapter}
\endbibitem

\bibitem[\protect\citeauthoryear{Dosovitskiy et~al.}{2021}]{dosovitskiy2021an}
\begin{bchapter}
\bauthor{\bsnm{Dosovitskiy}, \binits{A.}},
\bauthor{\bsnm{Beyer}, \binits{L.}},
\bauthor{\bsnm{Kolesnikov}, \binits{A.}},
\bauthor{\bsnm{Weissenborn}, \binits{D.}},
\bauthor{\bsnm{Zhai}, \binits{X.}},
\bauthor{\bsnm{Unterthiner}, \binits{T.}},
\bauthor{\bsnm{Dehghani}, \binits{M.}},
\bauthor{\bsnm{Minderer}, \binits{M.}},
\bauthor{\bsnm{Heigold}, \binits{G.}},
\bauthor{\bsnm{Gelly}, \binits{S.}},
\bauthor{\bsnm{Uszkoreit}, \binits{J.}},
\bauthor{\bsnm{Houlsby}, \binits{N.}}:
\bctitle{An image is worth 16x16 words: Transformers for image recognition at scale}.
In: \bbtitle{Proceedings of the International Conference on Learning Representations},
pp. \bfpage{1}--\blpage{21}
(\byear{2021})
\end{bchapter}
\endbibitem

\bibitem[\protect\citeauthoryear{Radford et~al.}{2021}]{radford2021learning}
\begin{bchapter}
\bauthor{\bsnm{Radford}, \binits{A.}},
\bauthor{\bsnm{Kim}, \binits{J.W.}},
\bauthor{\bsnm{Hallacy}, \binits{C.}},
\bauthor{\bsnm{Ramesh}, \binits{A.}},
\bauthor{\bsnm{Goh}, \binits{G.}},
\bauthor{\bsnm{Agarwal}, \binits{S.}},
\bauthor{\bsnm{Sastry}, \binits{G.}},
\bauthor{\bsnm{Askell}, \binits{A.}},
\bauthor{\bsnm{Mishkin}, \binits{P.}},
\bauthor{\bsnm{Clark}, \binits{J.}}, \betal:
\bctitle{Learning transferable visual models from natural language supervision}.
In: \bbtitle{International Conference on Machine Learning},
pp. \bfpage{8748}--\blpage{8763}
(\byear{2021}).
\bcomment{PMLR}
\end{bchapter}
\endbibitem

\bibitem[\protect\citeauthoryear{Girdhar et~al.}{2023}]{girdhar2023imagebind}
\begin{bchapter}
\bauthor{\bsnm{Girdhar}, \binits{R.}},
\bauthor{\bsnm{El-Nouby}, \binits{A.}},
\bauthor{\bsnm{Liu}, \binits{Z.}},
\bauthor{\bsnm{Singh}, \binits{M.}},
\bauthor{\bsnm{Alwala}, \binits{K.V.}},
\bauthor{\bsnm{Joulin}, \binits{A.}},
\bauthor{\bsnm{Misra}, \binits{I.}}:
\bctitle{Imagebind: One embedding space to bind them all}.
In: \bbtitle{Proceedings of the IEEE/CVF Conference on Computer Vision and Pattern Recognition},
pp. \bfpage{15180}--\blpage{15190}
(\byear{2023})
\end{bchapter}
\endbibitem

\bibitem[\protect\citeauthoryear{Liu et~al.}{2023}]{liu2023visual}
\begin{bchapter}
\bauthor{\bsnm{Liu}, \binits{H.}},
\bauthor{\bsnm{Li}, \binits{C.}},
\bauthor{\bsnm{Wu}, \binits{Q.}},
\bauthor{\bsnm{Lee}, \binits{Y.J.}}:
\bctitle{Visual instruction tuning}.
In: \bbtitle{Proceedings of the Thirty-seventh Conference on Neural Information Processing Systems}
(\byear{2023})
\end{bchapter}
\endbibitem

\bibitem[\protect\citeauthoryear{Oquab et~al.}{2024}]{oquab2024dinov}
\begin{botherref}
\oauthor{\bsnm{Oquab}, \binits{M.}},
\oauthor{\bsnm{Darcet}, \binits{T.}},
\oauthor{\bsnm{Moutakanni}, \binits{T.}},
\oauthor{\bsnm{Vo}, \binits{H.V.}},
\oauthor{\bsnm{Szafraniec}, \binits{M.}},
\oauthor{\bsnm{Khalidov}, \binits{V.}},
\oauthor{\bsnm{Fernandez}, \binits{P.}},
\oauthor{\bsnm{HAZIZA}, \binits{D.}},
\oauthor{\bsnm{Massa}, \binits{F.}},
\oauthor{\bsnm{El-Nouby}, \binits{A.}},
\oauthor{\bsnm{Assran}, \binits{M.}},
\oauthor{\bsnm{Ballas}, \binits{N.}},
\oauthor{\bsnm{Galuba}, \binits{W.}},
\oauthor{\bsnm{Howes}, \binits{R.}},
\oauthor{\bsnm{Huang}, \binits{P.-Y.}},
\oauthor{\bsnm{Li}, \binits{S.-W.}},
\oauthor{\bsnm{Misra}, \binits{I.}},
\oauthor{\bsnm{Rabbat}, \binits{M.}},
\oauthor{\bsnm{Sharma}, \binits{V.}},
\oauthor{\bsnm{Synnaeve}, \binits{G.}},
\oauthor{\bsnm{Xu}, \binits{H.}},
\oauthor{\bsnm{Jegou}, \binits{H.}},
\oauthor{\bsnm{Mairal}, \binits{J.}},
\oauthor{\bsnm{Labatut}, \binits{P.}},
\oauthor{\bsnm{Joulin}, \binits{A.}},
\oauthor{\bsnm{Bojanowski}, \binits{P.}}:
{DINO}v2: Learning robust visual features without supervision.
Transactions on Machine Learning Research
(2024)
\end{botherref}
\endbibitem

\bibitem[\protect\citeauthoryear{Luo et~al.}{2025}]{luo2025deem}
\begin{bchapter}
\bauthor{\bsnm{Luo}, \binits{R.}},
\bauthor{\bsnm{Li}, \binits{Y.}},
\bauthor{\bsnm{Chen}, \binits{L.}},
\bauthor{\bsnm{He}, \binits{W.}},
\bauthor{\bsnm{Lin}, \binits{T.-E.}},
\bauthor{\bsnm{Liu}, \binits{Z.}},
\bauthor{\bsnm{Zhang}, \binits{L.}},
\bauthor{\bsnm{Song}, \binits{Z.}},
\bauthor{\bsnm{Rokny}, \binits{H.}},
\bauthor{\bsnm{Xia}, \binits{X.}},
\bauthor{\bsnm{Liu}, \binits{T.}},
\bauthor{\bsnm{Hui}, \binits{B.}},
\bauthor{\bsnm{Yang}, \binits{M.}}:
\bctitle{{DEEM}: Diffusion models serve as the eyes of large language models for image perception}.
In: \bbtitle{Proceedings of the Thirteenth International Conference on Learning Representations}
(\byear{2025})
\end{bchapter}
\endbibitem

\bibitem[\protect\citeauthoryear{Zeiler and Fergus}{2014}]{zeiler2014visualizing}
\begin{bchapter}
\bauthor{\bsnm{Zeiler}, \binits{M.D.}},
\bauthor{\bsnm{Fergus}, \binits{R.}}:
\bctitle{Visualizing and understanding convolutional networks}.
In: \bbtitle{Proceedings of the 13th European Computer Vision Conference},
pp. \bfpage{818}--\blpage{833}
(\byear{2014}).
\bcomment{Springer}
\end{bchapter}
\endbibitem

\bibitem[\protect\citeauthoryear{Szegedy et~al.}{2015}]{szegedy2015going}
\begin{bchapter}
\bauthor{\bsnm{Szegedy}, \binits{C.}},
\bauthor{\bsnm{Liu}, \binits{W.}},
\bauthor{\bsnm{Jia}, \binits{Y.}},
\bauthor{\bsnm{Sermanet}, \binits{P.}},
\bauthor{\bsnm{Reed}, \binits{S.}},
\bauthor{\bsnm{Anguelov}, \binits{D.}},
\bauthor{\bsnm{Erhan}, \binits{D.}},
\bauthor{\bsnm{Vanhoucke}, \binits{V.}},
\bauthor{\bsnm{Rabinovich}, \binits{A.}}:
\bctitle{Going deeper with convolutions}.
In: \bbtitle{Proceedings of the IEEE Conference on Computer Vision and Pattern Recognition},
pp. \bfpage{1}--\blpage{9}
(\byear{2015})
\end{bchapter}
\endbibitem

\bibitem[\protect\citeauthoryear{Howard et~al.}{2019}]{howard2019searching}
\begin{bchapter}
\bauthor{\bsnm{Howard}, \binits{A.}},
\bauthor{\bsnm{Sandler}, \binits{M.}},
\bauthor{\bsnm{Chu}, \binits{G.}},
\bauthor{\bsnm{Chen}, \binits{L.-C.}},
\bauthor{\bsnm{Chen}, \binits{B.}},
\bauthor{\bsnm{Tan}, \binits{M.}},
\bauthor{\bsnm{Wang}, \binits{W.}},
\bauthor{\bsnm{Zhu}, \binits{Y.}},
\bauthor{\bsnm{Pang}, \binits{R.}},
\bauthor{\bsnm{Vasudevan}, \binits{V.}}, \betal:
\bctitle{Searching for mobilenetv3}.
In: \bbtitle{Proceedings of the IEEE/CVF International Conference on Computer Vision},
pp. \bfpage{1314}--\blpage{1324}
(\byear{2019})
\end{bchapter}
\endbibitem

\bibitem[\protect\citeauthoryear{Tan and Le}{2019}]{tan2019efficientnet}
\begin{bchapter}
\bauthor{\bsnm{Tan}, \binits{M.}},
\bauthor{\bsnm{Le}, \binits{Q.}}:
\bctitle{Efficientnet: Rethinking model scaling for convolutional neural networks}.
In: \bbtitle{Proceedings of the International Conference on Machine Learning},
pp. \bfpage{6105}--\blpage{6114}
(\byear{2019}).
\bcomment{PMLR}
\end{bchapter}
\endbibitem

\bibitem[\protect\citeauthoryear{Carion et~al.}{2020}]{carion2020end}
\begin{bchapter}
\bauthor{\bsnm{Carion}, \binits{N.}},
\bauthor{\bsnm{Massa}, \binits{F.}},
\bauthor{\bsnm{Synnaeve}, \binits{G.}},
\bauthor{\bsnm{Usunier}, \binits{N.}},
\bauthor{\bsnm{Kirillov}, \binits{A.}},
\bauthor{\bsnm{Zagoruyko}, \binits{S.}}:
\bctitle{End-to-end object detection with transformers}.
In: \bbtitle{European Conference on Computer Vision},
pp. \bfpage{213}--\blpage{229}
(\byear{2020}).
\bcomment{Springer}
\end{bchapter}
\endbibitem

\bibitem[\protect\citeauthoryear{Cheng et~al.}{2022}]{cheng2022masked}
\begin{bchapter}
\bauthor{\bsnm{Cheng}, \binits{B.}},
\bauthor{\bsnm{Misra}, \binits{I.}},
\bauthor{\bsnm{Schwing}, \binits{A.G.}},
\bauthor{\bsnm{Kirillov}, \binits{A.}},
\bauthor{\bsnm{Girdhar}, \binits{R.}}:
\bctitle{Masked-attention mask transformer for universal image segmentation}.
In: \bbtitle{Proceedings of the IEEE/CVF Conference on Computer Vision and Pattern Recognition},
pp. \bfpage{1290}--\blpage{1299}
(\byear{2022})
\end{bchapter}
\endbibitem

\bibitem[\protect\citeauthoryear{Liu et~al.}{2022}]{liu2022convnet}
\begin{bchapter}
\bauthor{\bsnm{Liu}, \binits{Z.}},
\bauthor{\bsnm{Mao}, \binits{H.}},
\bauthor{\bsnm{Wu}, \binits{C.-Y.}},
\bauthor{\bsnm{Feichtenhofer}, \binits{C.}},
\bauthor{\bsnm{Darrell}, \binits{T.}},
\bauthor{\bsnm{Xie}, \binits{S.}}:
\bctitle{A convnet for the 2020s}.
In: \bbtitle{Proceedings of the IEEE/CVF Conference on Computer Vision and Pattern Recognition},
pp. \bfpage{11976}--\blpage{11986}
(\byear{2022})
\end{bchapter}
\endbibitem

\bibitem[\protect\citeauthoryear{Kirillov et~al.}{2023}]{kirillov2023segment}
\begin{bchapter}
\bauthor{\bsnm{Kirillov}, \binits{A.}},
\bauthor{\bsnm{Mintun}, \binits{E.}},
\bauthor{\bsnm{Ravi}, \binits{N.}},
\bauthor{\bsnm{Mao}, \binits{H.}},
\bauthor{\bsnm{Rolland}, \binits{C.}},
\bauthor{\bsnm{Gustafson}, \binits{L.}},
\bauthor{\bsnm{Xiao}, \binits{T.}},
\bauthor{\bsnm{Whitehead}, \binits{S.}},
\bauthor{\bsnm{Berg}, \binits{A.C.}},
\bauthor{\bsnm{Lo}, \binits{W.-Y.}}, \betal:
\bctitle{Segment anything}.
In: \bbtitle{Proceedings of the IEEE/CVF International Conference on Computer Vision},
pp. \bfpage{4015}--\blpage{4026}
(\byear{2023})
\end{bchapter}
\endbibitem

\bibitem[\protect\citeauthoryear{Gupta et~al.}{2025}]{gupta2025evaluating}
\begin{bchapter}
\bauthor{\bsnm{Gupta}, \binits{S.}},
\bauthor{\bsnm{Kumar}, \binits{R.}},
\bauthor{\bsnm{Raja}, \binits{K.}},
\bauthor{\bsnm{Crispo}, \binits{B.}},
\bauthor{\bsnm{Maple}, \binits{C.}}:
\bctitle{Evaluating a bimodal user verification robustness against synthetic data attacks}.
In: \bbtitle{Proceeding of the International Conference on Security and Cryptography (SECRYPT)},
pp. \bfpage{1}--\blpage{12}
(\byear{2025})
\end{bchapter}
\endbibitem

\bibitem[\protect\citeauthoryear{Shu et~al.}{2021}]{shu2021encoding}
\begin{barticle}
\bauthor{\bsnm{Shu}, \binits{M.}},
\bauthor{\bsnm{Wu}, \binits{Z.}},
\bauthor{\bsnm{Goldblum}, \binits{M.}},
\bauthor{\bsnm{Goldstein}, \binits{T.}}:
\batitle{Encoding robustness to image style via adversarial feature perturbations}.
\bjtitle{Advances in Neural Information Processing Systems}
\bvolume{34},
\bfpage{28042}--\blpage{28053}
(\byear{2021})
\end{barticle}
\endbibitem

\bibitem[\protect\citeauthoryear{Zhu et~al.}{2023}]{zhu2023understanding}
\begin{bchapter}
\bauthor{\bsnm{Zhu}, \binits{Z.}},
\bauthor{\bsnm{Zhang}, \binits{Y.}},
\bauthor{\bsnm{Chen}, \binits{H.}},
\bauthor{\bsnm{Dong}, \binits{Y.}},
\bauthor{\bsnm{Zhao}, \binits{S.}},
\bauthor{\bsnm{Ding}, \binits{W.}},
\bauthor{\bsnm{Zhong}, \binits{J.}},
\bauthor{\bsnm{Zheng}, \binits{S.}}:
\bctitle{Understanding the robustness of 3d object detection with bird's-eye-view representations in autonomous driving}.
In: \bbtitle{Proceedings of the IEEE/CVF Conference on Computer Vision and Pattern Recognition},
pp. \bfpage{21600}--\blpage{21610}
(\byear{2023})
\end{bchapter}
\endbibitem

\bibitem[\protect\citeauthoryear{Papernot et~al.}{2016}]{papernot2016distillation}
\begin{bchapter}
\bauthor{\bsnm{Papernot}, \binits{N.}},
\bauthor{\bsnm{McDaniel}, \binits{P.}},
\bauthor{\bsnm{Wu}, \binits{X.}},
\bauthor{\bsnm{Jha}, \binits{S.}},
\bauthor{\bsnm{Swami}, \binits{A.}}:
\bctitle{Distillation as a defense to adversarial perturbations against deep neural networks}.
In: \bbtitle{Proceedings of the IEEE Symposium on Security and Privacy (SP)},
pp. \bfpage{582}--\blpage{597}
(\byear{2016}).
\bcomment{IEEE}
\end{bchapter}
\endbibitem

\bibitem[\protect\citeauthoryear{Gardner et~al.}{2024}]{gardner2024benchmarking}
\begin{botherref}
\oauthor{\bsnm{Gardner}, \binits{J.}},
\oauthor{\bsnm{Popovic}, \binits{Z.}},
\oauthor{\bsnm{Schmidt}, \binits{L.}}:
Benchmarking distribution shift in tabular data with tableshift.
Advances in Neural Information Processing Systems
\textbf{36}
(2024)
\end{botherref}
\endbibitem

\bibitem[\protect\citeauthoryear{Feng et~al.}{2023}]{feng2023robust}
\begin{barticle}
\bauthor{\bsnm{Feng}, \binits{W.}},
\bauthor{\bsnm{Xu}, \binits{N.}},
\bauthor{\bsnm{Zhang}, \binits{T.}},
\bauthor{\bsnm{Wu}, \binits{B.}},
\bauthor{\bsnm{Zhang}, \binits{Y.}}:
\batitle{Robust and generalized physical adversarial attacks via meta-gan}.
\bjtitle{IEEE Transactions on Information Forensics and Security}
\bvolume{19},
\bfpage{1112}--\blpage{1125}
(\byear{2023})
\end{barticle}
\endbibitem

\bibitem[\protect\citeauthoryear{Bitton et~al.}{2023}]{bitton2023evaluating}
\begin{barticle}
\bauthor{\bsnm{Bitton}, \binits{R.}},
\bauthor{\bsnm{Maman}, \binits{N.}},
\bauthor{\bsnm{Singh}, \binits{I.}},
\bauthor{\bsnm{Momiyama}, \binits{S.}},
\bauthor{\bsnm{Elovici}, \binits{Y.}},
\bauthor{\bsnm{Shabtai}, \binits{A.}}:
\batitle{Evaluating the cybersecurity risk of real-world, machine learning production systems}.
\bjtitle{ACM Computing Surveys}
\bvolume{55}(\bissue{9}),
\bfpage{1}--\blpage{36}
(\byear{2023})
\end{barticle}
\endbibitem

\bibitem[\protect\citeauthoryear{Meyers et~al.}{2023}]{meyers2023safety}
\begin{barticle}
\bauthor{\bsnm{Meyers}, \binits{C.}},
\bauthor{\bsnm{L{\"o}fstedt}, \binits{T.}},
\bauthor{\bsnm{Elmroth}, \binits{E.}}:
\batitle{Safety-critical computer vision: an empirical survey of adversarial evasion attacks and defenses on computer vision systems}.
\bjtitle{Artificial Intelligence Review}
\bvolume{56}(\bissue{Suppl 1}),
\bfpage{217}--\blpage{251}
(\byear{2023})
\end{barticle}
\endbibitem

\bibitem[\protect\citeauthoryear{Xu et~al.}{2024}]{xu2024demystifying}
\begin{bchapter}
\bauthor{\bsnm{Xu}, \binits{H.}},
\bauthor{\bsnm{Xie}, \binits{S.}},
\bauthor{\bsnm{Tan}, \binits{X.}},
\bauthor{\bsnm{Huang}, \binits{P.-Y.}},
\bauthor{\bsnm{Howes}, \binits{R.}},
\bauthor{\bsnm{Sharma}, \binits{V.}},
\bauthor{\bsnm{Li}, \binits{S.-W.}},
\bauthor{\bsnm{Ghosh}, \binits{G.}},
\bauthor{\bsnm{Zettlemoyer}, \binits{L.}},
\bauthor{\bsnm{Feichtenhofer}, \binits{C.}}:
\bctitle{Demystifying {CLIP} data}.
In: \bbtitle{Proceedings of the Twelfth International Conference on Learning Representations}
(\byear{2024})
\end{bchapter}
\endbibitem

\bibitem[\protect\citeauthoryear{Li et~al.}{2021}]{li2021survey}
\begin{barticle}
\bauthor{\bsnm{Li}, \binits{Z.}},
\bauthor{\bsnm{Liu}, \binits{F.}},
\bauthor{\bsnm{Yang}, \binits{W.}},
\bauthor{\bsnm{Peng}, \binits{S.}},
\bauthor{\bsnm{Zhou}, \binits{J.}}:
\batitle{A survey of convolutional neural networks: analysis, applications, and prospects}.
\bjtitle{IEEE transactions on neural networks and learning systems}
\bvolume{33}(\bissue{12}),
\bfpage{6999}--\blpage{7019}
(\byear{2021})
\end{barticle}
\endbibitem

\bibitem[\protect\citeauthoryear{Zhao et~al.}{2023}]{zhao2023evaluating}
\begin{barticle}
\bauthor{\bsnm{Zhao}, \binits{Y.}},
\bauthor{\bsnm{Pang}, \binits{T.}},
\bauthor{\bsnm{Du}, \binits{C.}},
\bauthor{\bsnm{Yang}, \binits{X.}},
\bauthor{\bsnm{Li}, \binits{C.}},
\bauthor{\bsnm{Cheung}, \binits{N.-M.M.}},
\bauthor{\bsnm{Lin}, \binits{M.}}:
\batitle{On evaluating adversarial robustness of large vision-language models}.
\bjtitle{Advances in Neural Information Processing Systems}
\bvolume{36},
\bfpage{54111}--\blpage{54138}
(\byear{2023})
\end{barticle}
\endbibitem

\bibitem[\protect\citeauthoryear{Badrinarayanan et~al.}{2017}]{badrinarayanan2017segnet}
\begin{barticle}
\bauthor{\bsnm{Badrinarayanan}, \binits{V.}},
\bauthor{\bsnm{Kendall}, \binits{A.}},
\bauthor{\bsnm{Cipolla}, \binits{R.}}:
\batitle{Segnet: A deep convolutional encoder-decoder architecture for image segmentation}.
\bjtitle{IEEE transactions on pattern analysis and machine intelligence}
\bvolume{39}(\bissue{12}),
\bfpage{2481}--\blpage{2495}
(\byear{2017})
\end{barticle}
\endbibitem

\bibitem[\protect\citeauthoryear{Usman et~al.}{2024}]{usman2024enhanced}
\begin{barticle}
\bauthor{\bsnm{Usman}, \binits{M.}},
\bauthor{\bsnm{Zaka-Ud-Din}, \binits{M.}},
\bauthor{\bsnm{Ling}, \binits{Q.}}:
\batitle{Enhanced encoder--decoder architecture for visual perception multitasking of autonomous driving}.
\bjtitle{Expert Systems with Applications}
\bvolume{246},
\bfpage{123249}
(\byear{2024})
\end{barticle}
\endbibitem

\bibitem[\protect\citeauthoryear{Gao et~al.}{2021}]{gao2021simcse}
\begin{bchapter}
\bauthor{\bsnm{Gao}, \binits{T.}},
\bauthor{\bsnm{Yao}, \binits{X.}},
\bauthor{\bsnm{Chen}, \binits{D.}}:
\bctitle{{SimCSE}: Simple contrastive learning of sentence embeddings}.
In: \bbtitle{Proceedings of the Empirical Methods in Natural Language Processing (EMNLP)},
pp. \bfpage{6894}--\blpage{6910}
(\byear{2021})
\end{bchapter}
\endbibitem

\bibitem[\protect\citeauthoryear{Tziafas and Kasaei}{2023}]{tziafas2023early}
\begin{bchapter}
\bauthor{\bsnm{Tziafas}, \binits{G.}},
\bauthor{\bsnm{Kasaei}, \binits{H.}}:
\bctitle{Early or late fusion matters: Efficient rgb-d fusion in vision transformers for 3d object recognition}.
In: \bbtitle{Proceeding of the International Conference on Intelligent Robots and Systems (IROS)},
pp. \bfpage{9558}--\blpage{9565}
(\byear{2023}).
\bcomment{IEEE}
\end{bchapter}
\endbibitem

\bibitem[\protect\citeauthoryear{Huang et~al.}{2024}]{huang2024early}
\begin{bchapter}
\bauthor{\bsnm{Huang}, \binits{H.}},
\bauthor{\bsnm{Liu}, \binits{F.}},
\bauthor{\bsnm{Fu}, \binits{L.}},
\bauthor{\bsnm{Wu}, \binits{T.}},
\bauthor{\bsnm{Mukadam}, \binits{M.}},
\bauthor{\bsnm{Malik}, \binits{J.}},
\bauthor{\bsnm{Goldberg}, \binits{K.}},
\bauthor{\bsnm{Abbeel}, \binits{P.}}:
\bctitle{Early fusion helps vision language action models generalize better}.
In: \bbtitle{Proceeding of the 1st Workshop on X-Embodiment Robot Learning},
pp. \bfpage{1}--\blpage{15}
(\byear{2024})
\end{bchapter}
\endbibitem

\bibitem[\protect\citeauthoryear{Dong et~al.}{2022}]{dong2022exploring}
\begin{bchapter}
\bauthor{\bsnm{Dong}, \binits{Z.}},
\bauthor{\bsnm{Ni}, \binits{J.}},
\bauthor{\bsnm{Bikel}, \binits{D.}},
\bauthor{\bsnm{Alfonseca}, \binits{E.}},
\bauthor{\bsnm{Wang}, \binits{Y.}},
\bauthor{\bsnm{Qu}, \binits{C.}},
\bauthor{\bsnm{Zitouni}, \binits{I.}}:
\bctitle{Exploring dual encoder architectures for question answering}.
In: \bbtitle{Proceedings of the 2022 Conference on Empirical Methods in Natural Language Processing},
pp. \bfpage{9414}--\blpage{9419}.
\bpublisher{Association for Computational Linguistics}, \blocation{???}
(\byear{2022})
\end{bchapter}
\endbibitem

\bibitem[\protect\citeauthoryear{Li et~al.}{2023}]{li2023trustworthy}
\begin{barticle}
\bauthor{\bsnm{Li}, \binits{B.}},
\bauthor{\bsnm{Qi}, \binits{P.}},
\bauthor{\bsnm{Liu}, \binits{B.}},
\bauthor{\bsnm{Di}, \binits{S.}},
\bauthor{\bsnm{Liu}, \binits{J.}},
\bauthor{\bsnm{Pei}, \binits{J.}},
\bauthor{\bsnm{Yi}, \binits{J.}},
\bauthor{\bsnm{Zhou}, \binits{B.}}:
\batitle{Trustworthy ai: From principles to practices}.
\bjtitle{ACM Computing Surveys}
\bvolume{55}(\bissue{9}),
\bfpage{1}--\blpage{46}
(\byear{2023})
\end{barticle}
\endbibitem

\bibitem[\protect\citeauthoryear{Bhojanapalli et~al.}{2021}]{bhojanapalli2021understanding}
\begin{bchapter}
\bauthor{\bsnm{Bhojanapalli}, \binits{S.}},
\bauthor{\bsnm{Chakrabarti}, \binits{A.}},
\bauthor{\bsnm{Glasner}, \binits{D.}},
\bauthor{\bsnm{Li}, \binits{D.}},
\bauthor{\bsnm{Unterthiner}, \binits{T.}},
\bauthor{\bsnm{Veit}, \binits{A.}}:
\bctitle{Understanding robustness of transformers for image classification}.
In: \bbtitle{Proceedings of the IEEE/CVF International Conference on Computer Vision},
pp. \bfpage{10231}--\blpage{10241}
(\byear{2021})
\end{bchapter}
\endbibitem

\bibitem[\protect\citeauthoryear{Gu et~al.}{2022}]{gu2022evaluating}
\begin{bchapter}
\bauthor{\bsnm{Gu}, \binits{J.}},
\bauthor{\bsnm{Tresp}, \binits{V.}},
\bauthor{\bsnm{Qin}, \binits{Y.}}:
\bctitle{Evaluating model robustness to patch perturbations}.
In: \bbtitle{Proceeding of the Shift Happens Workshop (ICML)},
pp. \bfpage{1}--\blpage{6}
(\byear{2022})
\end{bchapter}
\endbibitem

\bibitem[\protect\citeauthoryear{Han et~al.}{2023}]{han2023interpreting}
\begin{botherref}
\oauthor{\bsnm{Han}, \binits{S.}},
\oauthor{\bsnm{Lin}, \binits{C.}},
\oauthor{\bsnm{Shen}, \binits{C.}},
\oauthor{\bsnm{Wang}, \binits{Q.}},
\oauthor{\bsnm{Guan}, \binits{X.}}:
Interpreting adversarial examples in deep learning: A review.
ACM Computing Surveys
(2023)
\end{botherref}
\endbibitem

\bibitem[\protect\citeauthoryear{Qian et~al.}{2022}]{qian2022survey}
\begin{barticle}
\bauthor{\bsnm{Qian}, \binits{Z.}},
\bauthor{\bsnm{Huang}, \binits{K.}},
\bauthor{\bsnm{Wang}, \binits{Q.-F.}},
\bauthor{\bsnm{Zhang}, \binits{X.-Y.}}:
\batitle{A survey of robust adversarial training in pattern recognition: Fundamental, theory, and methodologies}.
\bjtitle{Pattern Recognition}
\bvolume{131},
\bfpage{108889}
(\byear{2022})
\end{barticle}
\endbibitem

\bibitem[\protect\citeauthoryear{Hendrycks and Dietterich}{2018}]{hendrycks2018benchmarking}
\begin{bchapter}
\bauthor{\bsnm{Hendrycks}, \binits{D.}},
\bauthor{\bsnm{Dietterich}, \binits{T.}}:
\bctitle{Benchmarking neural network robustness to common corruptions and perturbations}.
In: \bbtitle{Proceedings of the International Conference on Learning Representations},
pp. \bfpage{1}--\blpage{16}
(\byear{2018})
\end{bchapter}
\endbibitem

\bibitem[\protect\citeauthoryear{Meng et~al.}{2022}]{meng2022adversarial}
\begin{botherref}
\oauthor{\bsnm{Meng}, \binits{M.H.}},
\oauthor{\bsnm{Bai}, \binits{G.}},
\oauthor{\bsnm{Teo}, \binits{S.G.}},
\oauthor{\bsnm{Hou}, \binits{Z.}},
\oauthor{\bsnm{Xiao}, \binits{Y.}},
\oauthor{\bsnm{Lin}, \binits{Y.}},
\oauthor{\bsnm{Dong}, \binits{J.S.}}:
Adversarial robustness of deep neural networks: A survey from a formal verification perspective.
IEEE Transactions on Dependable and Secure Computing
(2022)
\end{botherref}
\endbibitem

\bibitem[\protect\citeauthoryear{Fan et~al.}{2022}]{fan2022survey}
\begin{bchapter}
\bauthor{\bsnm{Fan}, \binits{J.}},
\bauthor{\bsnm{Yan}, \binits{Q.}},
\bauthor{\bsnm{Li}, \binits{M.}},
\bauthor{\bsnm{Qu}, \binits{G.}},
\bauthor{\bsnm{Xiao}, \binits{Y.}}:
\bctitle{A survey on data poisoning attacks and defenses}.
In: \bbtitle{Proceedings of the IEEE International Conference on Data Science in Cyberspace (DSC)},
pp. \bfpage{48}--\blpage{55}
(\byear{2022}).
\bcomment{IEEE}
\end{bchapter}
\endbibitem

\bibitem[\protect\citeauthoryear{Wang et~al.}{2023}]{wang2023evasion}
\begin{botherref}
\oauthor{\bsnm{Wang}, \binits{S.}},
\oauthor{\bsnm{Ko}, \binits{R.K.}},
\oauthor{\bsnm{Bai}, \binits{G.}},
\oauthor{\bsnm{Dong}, \binits{N.}},
\oauthor{\bsnm{Choi}, \binits{T.}},
\oauthor{\bsnm{Zhang}, \binits{Y.}}:
Evasion attack and defense on machine learning models in cyber-physical systems: A survey.
IEEE Communications Surveys \& Tutorials
(2023)
\end{botherref}
\endbibitem

\bibitem[\protect\citeauthoryear{Li et~al.}{2024}]{li2024survey}
\begin{barticle}
\bauthor{\bsnm{Li}, \binits{Y.}},
\bauthor{\bsnm{Xie}, \binits{B.}},
\bauthor{\bsnm{Guo}, \binits{S.}},
\bauthor{\bsnm{Yang}, \binits{Y.}},
\bauthor{\bsnm{Xiao}, \binits{B.}}:
\batitle{A survey of robustness and safety of 2d and 3d deep learning models against adversarial attacks}.
\bjtitle{ACM Computing Surveys}
\bvolume{56}(\bissue{6}),
\bfpage{1}--\blpage{37}
(\byear{2024})
\end{barticle}
\endbibitem

\bibitem[\protect\citeauthoryear{Ren et~al.}{2020}]{ren2020adversarial}
\begin{barticle}
\bauthor{\bsnm{Ren}, \binits{K.}},
\bauthor{\bsnm{Zheng}, \binits{T.}},
\bauthor{\bsnm{Qin}, \binits{Z.}},
\bauthor{\bsnm{Liu}, \binits{X.}}:
\batitle{Adversarial attacks and defenses in deep learning}.
\bjtitle{Engineering}
\bvolume{6}(\bissue{3}),
\bfpage{346}--\blpage{360}
(\byear{2020})
\end{barticle}
\endbibitem

\bibitem[\protect\citeauthoryear{Chen et~al.}{2024}]{chen2024learn}
\begin{botherref}
\oauthor{\bsnm{Chen}, \binits{Z.}},
\oauthor{\bsnm{Wang}, \binits{Z.}},
\oauthor{\bsnm{Xu}, \binits{D.}},
\oauthor{\bsnm{Zhu}, \binits{J.}},
\oauthor{\bsnm{Shen}, \binits{W.}},
\oauthor{\bsnm{Zheng}, \binits{S.}},
\oauthor{\bsnm{Xuan}, \binits{Q.}},
\oauthor{\bsnm{Yang}, \binits{X.}}:
Learn to defend: Adversarial multi-distillation for automatic modulation recognition models.
IEEE Transactions on Information Forensics and Security
(2024)
\end{botherref}
\endbibitem

\bibitem[\protect\citeauthoryear{Serban et~al.}{2020}]{serban2020adversarial}
\begin{barticle}
\bauthor{\bsnm{Serban}, \binits{A.}},
\bauthor{\bsnm{Poll}, \binits{E.}},
\bauthor{\bsnm{Visser}, \binits{J.}}:
\batitle{Adversarial examples on object recognition: A comprehensive survey}.
\bjtitle{ACM Computing Surveys (CSUR)}
\bvolume{53}(\bissue{3}),
\bfpage{1}--\blpage{38}
(\byear{2020})
\end{barticle}
\endbibitem

\bibitem[\protect\citeauthoryear{Zhao et~al.}{2023}]{zhao2023remix}
\begin{barticle}
\bauthor{\bsnm{Zhao}, \binits{H.}},
\bauthor{\bsnm{Hao}, \binits{L.}},
\bauthor{\bsnm{Hao}, \binits{K.}},
\bauthor{\bsnm{Wei}, \binits{B.}},
\bauthor{\bsnm{Cai}, \binits{X.}}:
\batitle{Remix: Towards the transferability of adversarial examples}.
\bjtitle{Neural Networks}
\bvolume{163},
\bfpage{367}--\blpage{378}
(\byear{2023})
\end{barticle}
\endbibitem

\bibitem[\protect\citeauthoryear{Pytorch}{2025a}]{pytorch2025vit}
\begin{botherref}
\oauthor{\bsnm{Pytorch}}:
ViT.
\url{https://docs.pytorch.org/vision/main/models/vision_transformer.html}.
online web resource
(2025)
\end{botherref}
\endbibitem

\bibitem[\protect\citeauthoryear{Pytorch}{2025b}]{pytorch2025resnet}
\begin{botherref}
\oauthor{\bsnm{Pytorch}}:
Resnet.
\url{https://pytorch.org/hub/pytorch_vision_resnet/}.
online web resource
(2025)
\end{botherref}
\endbibitem

\bibitem[\protect\citeauthoryear{Pytorch}{2025c}]{pytorch2024inception}
\begin{botherref}
\oauthor{\bsnm{Pytorch}}:
Inception V3.
\url{https://pytorch.org/hub/pytorch_vision_inception_v3/}.
online web resource
(2025)
\end{botherref}
\endbibitem

\bibitem[\protect\citeauthoryear{Kang et~al.}{2024}]{kang2024catch}
\begin{bchapter}
\bauthor{\bsnm{Kang}, \binits{M.}},
\bauthor{\bsnm{Kang}, \binits{M.}},
\bauthor{\bsnm{Kim}, \binits{S.}}:
\bctitle{Catch-up mix: Catch-up class for struggling filters in cnn}.
In: \bbtitle{Proceedings of the AAAI Conference on Artificial Intelligence},
vol. \bseriesno{38},
pp. \bfpage{2705}--\blpage{2713}
(\byear{2024})
\end{bchapter}
\endbibitem

\bibitem[\protect\citeauthoryear{Yang et~al.}{2024}]{yang2024generalized}
\begin{barticle}
\bauthor{\bsnm{Yang}, \binits{J.}},
\bauthor{\bsnm{Zhou}, \binits{K.}},
\bauthor{\bsnm{Li}, \binits{Y.}},
\bauthor{\bsnm{Liu}, \binits{Z.}}:
\batitle{Generalized out-of-distribution detection: A survey}.
\bjtitle{International Journal of Computer Vision}
\bvolume{132}(\bissue{12}),
\bfpage{5635}--\blpage{5662}
(\byear{2024})
\end{barticle}
\endbibitem

\bibitem[\protect\citeauthoryear{Chen et~al.}{2024}]{chen2024exploring}
\begin{barticle}
\bauthor{\bsnm{Chen}, \binits{Q.}},
\bauthor{\bsnm{Li}, \binits{K.}},
\bauthor{\bsnm{Chen}, \binits{Z.}},
\bauthor{\bsnm{Maul}, \binits{T.}},
\bauthor{\bsnm{Yin}, \binits{J.}}:
\batitle{Exploring feature sparsity for out-of-distribution detection}.
\bjtitle{Scientific Reports}
\bvolume{14}(\bissue{1}),
\bfpage{28444}
(\byear{2024})
\end{barticle}
\endbibitem

\bibitem[\protect\citeauthoryear{Liu et~al.}{2024}]{liu2024comprehensive}
\begin{botherref}
\oauthor{\bsnm{Liu}, \binits{C.}},
\oauthor{\bsnm{Dong}, \binits{Y.}},
\oauthor{\bsnm{Xiang}, \binits{W.}},
\oauthor{\bsnm{Yang}, \binits{X.}},
\oauthor{\bsnm{Su}, \binits{H.}},
\oauthor{\bsnm{Zhu}, \binits{J.}},
\oauthor{\bsnm{Chen}, \binits{Y.}},
\oauthor{\bsnm{He}, \binits{Y.}},
\oauthor{\bsnm{Xue}, \binits{H.}},
\oauthor{\bsnm{Zheng}, \binits{S.}}:
A comprehensive study on robustness of image classification models: Benchmarking and rethinking.
International Journal of Computer Vision,
1--23
(2024)
\end{botherref}
\endbibitem

\bibitem[\protect\citeauthoryear{Goodfellow et~al.}{2015}]{goodfellow2015explaining}
\begin{bchapter}
\bauthor{\bsnm{Goodfellow}, \binits{I.J.}},
\bauthor{\bsnm{Shlens}, \binits{J.}},
\bauthor{\bsnm{Szegedy}, \binits{C.}}:
\bctitle{Explaining and harnessing adversarial examples}.
In: \bbtitle{Proceedings of the 3rd International Conference on Learning Representations},
pp. \bfpage{1}--\blpage{10}
(\byear{2015})
\end{bchapter}
\endbibitem

\bibitem[\protect\citeauthoryear{Madry et~al.}{2018}]{madry2018towards}
\begin{bchapter}
\bauthor{\bsnm{Madry}, \binits{A.}},
\bauthor{\bsnm{Makelov}, \binits{A.}},
\bauthor{\bsnm{Schmidt}, \binits{L.}},
\bauthor{\bsnm{Tsipras}, \binits{D.}},
\bauthor{\bsnm{Vladu}, \binits{A.}}:
\bctitle{Towards deep learning models resistant to adversarial attacks}.
In: \bbtitle{Proceedings of the International Conference on Learning Representations},
pp. \bfpage{1}--\blpage{27}
(\byear{2018})
\end{bchapter}
\endbibitem

\bibitem[\protect\citeauthoryear{Carlini and Wagner}{2017}]{carlini2017towards}
\begin{bchapter}
\bauthor{\bsnm{Carlini}, \binits{N.}},
\bauthor{\bsnm{Wagner}, \binits{D.}}:
\bctitle{Towards evaluating the robustness of neural networks}.
In: \bbtitle{Proceedings of the IEEE Symposium on Security and Privacy (SP)},
pp. \bfpage{39}--\blpage{57}
(\byear{2017}).
\bcomment{Ieee}
\end{bchapter}
\endbibitem

\bibitem[\protect\citeauthoryear{Szegedy et~al.}{2014}]{szegedy2014intriguing}
\begin{bchapter}
\bauthor{\bsnm{Szegedy}, \binits{C.}},
\bauthor{\bsnm{Zare}, \binits{W.}},
\bauthor{\bsnm{Sutskever}, \binits{I.}},
\bauthor{\bsnm{Bruna}, \binits{J.}},
\bauthor{\bsnm{Erhan}, \binits{D.}},
\bauthor{\bsnm{Goodfellow}, \binits{I.}},
\bauthor{\bsnm{Fergus}, \binits{R.}}:
\bctitle{Intriguing properties of neural networks}.
In: \bbtitle{Proceedings of the International Conference on Learning Representations (ICLR)},
pp. \bfpage{1}--\blpage{10}
(\byear{2014})
\end{bchapter}
\endbibitem

\bibitem[\protect\citeauthoryear{Athalye et~al.}{2018}]{athalye2018obfuscated}
\begin{bchapter}
\bauthor{\bsnm{Athalye}, \binits{A.}},
\bauthor{\bsnm{Carlini}, \binits{N.}},
\bauthor{\bsnm{Wagner}, \binits{D.}}:
\bctitle{Obfuscated gradients give a false sense of security: Circumventing defenses to adversarial examples}.
In: \bbtitle{Proceedings of the International Conference on Machine Learning},
pp. \bfpage{274}--\blpage{283}
(\byear{2018}).
\bcomment{PMLR}
\end{bchapter}
\endbibitem

\bibitem[\protect\citeauthoryear{Shapira et~al.}{2023}]{shapira2023deep}
\begin{barticle}
\bauthor{\bsnm{Shapira}, \binits{Y.}},
\bauthor{\bsnm{Avneri}, \binits{E.}},
\bauthor{\bsnm{Drachsler-Cohen}, \binits{D.}}:
\batitle{Deep learning robustness verification for few-pixel attacks}.
\bjtitle{Proceedings of the ACM on Programming Languages}
\bvolume{7}(\bissue{OOPSLA1}),
\bfpage{434}--\blpage{461}
(\byear{2023})
\end{barticle}
\endbibitem

\bibitem[\protect\citeauthoryear{Zhu et~al.}{2023}]{zhu2023zeroth}
\begin{bchapter}
\bauthor{\bsnm{Zhu}, \binits{Y.}},
\bauthor{\bsnm{Zhao}, \binits{Y.}},
\bauthor{\bsnm{Hu}, \binits{Z.}},
\bauthor{\bsnm{Liu}, \binits{X.}},
\bauthor{\bsnm{Yan}, \binits{A.}}:
\bctitle{Zeroth-order gradient approximation based dast for black-box adversarial attacks}.
In: \bbtitle{Proceedings of the International Conference on Intelligent Computing},
pp. \bfpage{442}--\blpage{453}
(\byear{2023}).
\bcomment{Springer}
\end{bchapter}
\endbibitem

\bibitem[\protect\citeauthoryear{Jia et~al.}{2024}]{jia2024fooling}
\begin{botherref}
\oauthor{\bsnm{Jia}, \binits{W.}},
\oauthor{\bsnm{Lu}, \binits{Z.}},
\oauthor{\bsnm{Yu}, \binits{R.}},
\oauthor{\bsnm{Li}, \binits{L.}},
\oauthor{\bsnm{Zhang}, \binits{H.}},
\oauthor{\bsnm{Liu}, \binits{Z.}},
\oauthor{\bsnm{Qu}, \binits{G.}}:
Fooling decision-based black-box automotive vision perception systems in physical world.
IEEE Transactions on Intelligent Transportation Systems
(2024)
\end{botherref}
\endbibitem

\bibitem[\protect\citeauthoryear{Zhu et~al.}{2024}]{zhu2024review}
\begin{botherref}
\oauthor{\bsnm{Zhu}, \binits{Y.}},
\oauthor{\bsnm{Zhao}, \binits{Y.}},
\oauthor{\bsnm{Hu}, \binits{Z.}},
\oauthor{\bsnm{Luo}, \binits{T.}},
\oauthor{\bsnm{He}, \binits{L.}}:
A review of black-box adversarial attacks on image classification.
Neurocomputing,
128512
(2024)
\end{botherref}
\endbibitem

\bibitem[\protect\citeauthoryear{Tao et~al.}{2023}]{tao2023hard}
\begin{bchapter}
\bauthor{\bsnm{Tao}, \binits{G.}},
\bauthor{\bsnm{An}, \binits{S.}},
\bauthor{\bsnm{Cheng}, \binits{S.}},
\bauthor{\bsnm{Shen}, \binits{G.}},
\bauthor{\bsnm{Zhang}, \binits{X.}}:
\bctitle{Hard-label black-box universal adversarial patch attack}.
In: \bbtitle{Proceedings of the 32nd USENIX Security Symposium (USENIX Security 23)},
pp. \bfpage{697}--\blpage{714}
(\byear{2023})
\end{bchapter}
\endbibitem

\bibitem[\protect\citeauthoryear{Li et~al.}{2023}]{li2023sok}
\begin{bchapter}
\bauthor{\bsnm{Li}, \binits{L.}},
\bauthor{\bsnm{Xie}, \binits{T.}},
\bauthor{\bsnm{Li}, \binits{B.}}:
\bctitle{Sok: Certified robustness for deep neural networks}.
In: \bbtitle{Proceedings of the IEEE Symposium on Security and Privacy (SP)},
pp. \bfpage{1289}--\blpage{1310}
(\byear{2023}).
\bcomment{IEEE}
\end{bchapter}
\endbibitem

\bibitem[\protect\citeauthoryear{Yuan et~al.}{2019}]{yuan2019adversarial}
\begin{barticle}
\bauthor{\bsnm{Yuan}, \binits{X.}},
\bauthor{\bsnm{He}, \binits{P.}},
\bauthor{\bsnm{Zhu}, \binits{Q.}},
\bauthor{\bsnm{Li}, \binits{X.}}:
\batitle{Adversarial examples: Attacks and defenses for deep learning}.
\bjtitle{IEEE transactions on neural networks and learning systems}
\bvolume{30}(\bissue{9}),
\bfpage{2805}--\blpage{2824}
(\byear{2019})
\end{barticle}
\endbibitem

\bibitem[\protect\citeauthoryear{Chen et~al.}{2022}]{chen2022adversarial}
\begin{bchapter}
\bauthor{\bsnm{Chen}, \binits{Y.}},
\bauthor{\bsnm{Zhang}, \binits{M.}},
\bauthor{\bsnm{Li}, \binits{J.}},
\bauthor{\bsnm{Kuang}, \binits{X.}}:
\bctitle{Adversarial attacks and defenses in image classification: A practical perspective}.
In: \bbtitle{Proceedings of the 7th International Conference on Image, Vision and Computing (ICIVC)},
pp. \bfpage{424}--\blpage{430}
(\byear{2022}).
\bcomment{IEEE}
\end{bchapter}
\endbibitem

\bibitem[\protect\citeauthoryear{Li et~al.}{2024}]{li2024model}
\begin{barticle}
\bauthor{\bsnm{Li}, \binits{Q.}},
\bauthor{\bsnm{Chen}, \binits{J.}},
\bauthor{\bsnm{He}, \binits{K.}},
\bauthor{\bsnm{Zhang}, \binits{Z.}},
\bauthor{\bsnm{Du}, \binits{R.}},
\bauthor{\bsnm{She}, \binits{J.}},
\bauthor{\bsnm{Wang}, \binits{X.}}:
\batitle{Model-agnostic adversarial example detection via high-frequency amplification}.
\bjtitle{Computers \& Security}
\bvolume{141},
\bfpage{103791}
(\byear{2024})
\end{barticle}
\endbibitem

\bibitem[\protect\citeauthoryear{Pedraza et~al.}{2024}]{pedraza2024leveraging}
\begin{barticle}
\bauthor{\bsnm{Pedraza}, \binits{A.}},
\bauthor{\bsnm{Deniz}, \binits{O.}},
\bauthor{\bsnm{Singh}, \binits{H.}},
\bauthor{\bsnm{Bueno}, \binits{G.}}:
\batitle{Leveraging autoencoders and chaos theory to improve adversarial example detection}.
\bjtitle{Neural Computing and Applications}
\bvolume{36}(\bissue{29}),
\bfpage{18265}--\blpage{18275}
(\byear{2024})
\end{barticle}
\endbibitem

\bibitem[\protect\citeauthoryear{Wang et~al.}{2021}]{wang2021smsnet}
\begin{barticle}
\bauthor{\bsnm{Wang}, \binits{J.}},
\bauthor{\bsnm{Zhao}, \binits{J.}},
\bauthor{\bsnm{Yin}, \binits{Q.}},
\bauthor{\bsnm{Luo}, \binits{X.}},
\bauthor{\bsnm{Zheng}, \binits{Y.}},
\bauthor{\bsnm{Shi}, \binits{Y.-Q.}},
\bauthor{\bsnm{Jha}, \binits{S.K.}}:
\batitle{Smsnet: A new deep convolutional neural network model for adversarial example detection}.
\bjtitle{IEEE Transactions on Multimedia}
\bvolume{24},
\bfpage{230}--\blpage{244}
(\byear{2021})
\end{barticle}
\endbibitem

\bibitem[\protect\citeauthoryear{Luo et~al.}{2022}]{luo2022detecting}
\begin{barticle}
\bauthor{\bsnm{Luo}, \binits{W.}},
\bauthor{\bsnm{Wu}, \binits{C.}},
\bauthor{\bsnm{Ni}, \binits{L.}},
\bauthor{\bsnm{Zhou}, \binits{N.}},
\bauthor{\bsnm{Zhang}, \binits{Z.}}:
\batitle{Detecting adversarial examples by positive and negative representations}.
\bjtitle{Applied Soft Computing}
\bvolume{117},
\bfpage{108383}
(\byear{2022})
\end{barticle}
\endbibitem

\bibitem[\protect\citeauthoryear{Nesti et~al.}{2021}]{nesti2021detecting}
\begin{barticle}
\bauthor{\bsnm{Nesti}, \binits{F.}},
\bauthor{\bsnm{Biondi}, \binits{A.}},
\bauthor{\bsnm{Buttazzo}, \binits{G.}}:
\batitle{Detecting adversarial examples by input transformations, defense perturbations, and voting}.
\bjtitle{IEEE transactions on neural networks and learning systems}
\bvolume{34}(\bissue{3}),
\bfpage{1329}--\blpage{1341}
(\byear{2021})
\end{barticle}
\endbibitem

\bibitem[\protect\citeauthoryear{Xiong et~al.}{2022}]{xiong2022towards}
\begin{barticle}
\bauthor{\bsnm{Xiong}, \binits{P.}},
\bauthor{\bsnm{Buffett}, \binits{S.}},
\bauthor{\bsnm{Iqbal}, \binits{S.}},
\bauthor{\bsnm{Lamontagne}, \binits{P.}},
\bauthor{\bsnm{Mamun}, \binits{M.}},
\bauthor{\bsnm{Molyneaux}, \binits{H.}}:
\batitle{Towards a robust and trustworthy machine learning system development: An engineering perspective}.
\bjtitle{Journal of Information Security and Applications}
\bvolume{65},
\bfpage{103121}
(\byear{2022})
\end{barticle}
\endbibitem

\bibitem[\protect\citeauthoryear{Sun et~al.}{2019}]{sun2019adversarial}
\begin{bchapter}
\bauthor{\bsnm{Sun}, \binits{B.}},
\bauthor{\bsnm{Tsai}, \binits{N.-h.}},
\bauthor{\bsnm{Liu}, \binits{F.}},
\bauthor{\bsnm{Yu}, \binits{R.}},
\bauthor{\bsnm{Su}, \binits{H.}}:
\bctitle{Adversarial defense by stratified convolutional sparse coding}.
In: \bbtitle{Proceedings of the IEEE/CVF Conference on Computer Vision and Pattern Recognition},
pp. \bfpage{11447}--\blpage{11456}
(\byear{2019})
\end{bchapter}
\endbibitem

\bibitem[\protect\citeauthoryear{Xie et~al.}{2019}]{xie2019feature}
\begin{bchapter}
\bauthor{\bsnm{Xie}, \binits{C.}},
\bauthor{\bsnm{Wu}, \binits{Y.}},
\bauthor{\bsnm{Maaten}, \binits{L.v.d.}},
\bauthor{\bsnm{Yuille}, \binits{A.L.}},
\bauthor{\bsnm{He}, \binits{K.}}:
\bctitle{Feature denoising for improving adversarial robustness}.
In: \bbtitle{Proceedings of the IEEE/CVF Conference on Computer Vision and Pattern Recognition},
pp. \bfpage{501}--\blpage{509}
(\byear{2019})
\end{bchapter}
\endbibitem

\bibitem[\protect\citeauthoryear{Hu et~al.}{2024}]{hu2024efficient}
\begin{barticle}
\bauthor{\bsnm{Hu}, \binits{Y.}},
\bauthor{\bsnm{Tian}, \binits{C.}},
\bauthor{\bsnm{Zhang}, \binits{J.}},
\bauthor{\bsnm{Zhang}, \binits{S.}}:
\batitle{Efficient image denoising with heterogeneous kernel-based cnn}.
\bjtitle{Neurocomputing}
\bvolume{592},
\bfpage{127799}
(\byear{2024})
\end{barticle}
\endbibitem

\bibitem[\protect\citeauthoryear{Tiwari et~al.}{2022}]{tiwari2022regroup}
\begin{bchapter}
\bauthor{\bsnm{Tiwari}, \binits{L.}},
\bauthor{\bsnm{Madan}, \binits{A.}},
\bauthor{\bsnm{Anand}, \binits{S.}},
\bauthor{\bsnm{Banerjee}, \binits{S.}}:
\bctitle{Regroup: Rank-aggregating ensemble of generative classifiers for robust predictions}.
In: \bbtitle{Proceedings of the IEEE/CVF Winter Conference on Applications of Computer Vision},
pp. \bfpage{2595}--\blpage{2604}
(\byear{2022})
\end{bchapter}
\endbibitem

\bibitem[\protect\citeauthoryear{Wang et~al.}{2020}]{wang2020deep}
\begin{bchapter}
\bauthor{\bsnm{Wang}, \binits{L.}},
\bauthor{\bsnm{Zhang}, \binits{C.}},
\bauthor{\bsnm{Liu}, \binits{J.}}:
\bctitle{Deep learning defense method against adversarial attacks}.
In: \bbtitle{Proceedings of the International Conference on Systems, Man, and Cybernetics (SMC)},
pp. \bfpage{3667}--\blpage{3672}
(\byear{2020}).
\bcomment{IEEE}
\end{bchapter}
\endbibitem

\bibitem[\protect\citeauthoryear{Ma et~al.}{2024}]{ma2024adversarial}
\begin{botherref}
\oauthor{\bsnm{Ma}, \binits{Y.}},
\oauthor{\bsnm{Dong}, \binits{M.}},
\oauthor{\bsnm{Xu}, \binits{C.}}:
Adversarial robustness through random weight sampling.
Advances in Neural Information Processing Systems
\textbf{36}
(2024)
\end{botherref}
\endbibitem

\bibitem[\protect\citeauthoryear{Rice et~al.}{2020}]{rice2020overfitting}
\begin{bchapter}
\bauthor{\bsnm{Rice}, \binits{L.}},
\bauthor{\bsnm{Wong}, \binits{E.}},
\bauthor{\bsnm{Kolter}, \binits{Z.}}:
\bctitle{Overfitting in adversarially robust deep learning}.
In: \bbtitle{Proceeding of the International Conference on Machine Learning},
pp. \bfpage{8093}--\blpage{8104}
(\byear{2020}).
\bcomment{PMLR}
\end{bchapter}
\endbibitem

\bibitem[\protect\citeauthoryear{Jia et~al.}{2022}]{jia2022adversarial}
\begin{bchapter}
\bauthor{\bsnm{Jia}, \binits{X.}},
\bauthor{\bsnm{Zhang}, \binits{Y.}},
\bauthor{\bsnm{Wu}, \binits{B.}},
\bauthor{\bsnm{Ma}, \binits{K.}},
\bauthor{\bsnm{Wang}, \binits{J.}},
\bauthor{\bsnm{Cao}, \binits{X.}}:
\bctitle{Las-at: adversarial training with learnable attack strategy}.
In: \bbtitle{Proceedings of the IEEE/CVF Conference on Computer Vision and Pattern Recognition},
pp. \bfpage{13398}--\blpage{13408}
(\byear{2022})
\end{bchapter}
\endbibitem

\bibitem[\protect\citeauthoryear{Wei et~al.}{2023}]{wei2023cfa}
\begin{bchapter}
\bauthor{\bsnm{Wei}, \binits{Z.}},
\bauthor{\bsnm{Wang}, \binits{Y.}},
\bauthor{\bsnm{Guo}, \binits{Y.}},
\bauthor{\bsnm{Wang}, \binits{Y.}}:
\bctitle{Cfa: Class-wise calibrated fair adversarial training}.
In: \bbtitle{Proceedings of the IEEE/CVF Conference on Computer Vision and Pattern Recognition},
pp. \bfpage{8193}--\blpage{8201}
(\byear{2023})
\end{bchapter}
\endbibitem

\bibitem[\protect\citeauthoryear{Frosio and Kautz}{2023}]{frosio2023best}
\begin{bchapter}
\bauthor{\bsnm{Frosio}, \binits{I.}},
\bauthor{\bsnm{Kautz}, \binits{J.}}:
\bctitle{The best defense is a good offense: adversarial augmentation against adversarial attacks}.
In: \bbtitle{Proceedings of the IEEE/CVF Conference on Computer Vision and Pattern Recognition},
pp. \bfpage{4067}--\blpage{4076}
(\byear{2023})
\end{bchapter}
\endbibitem

\bibitem[\protect\citeauthoryear{Zhang}{2020}]{zhang2020machine}
\begin{botherref}
\oauthor{\bsnm{Zhang}, \binits{H.}}:
Machine learning with provable robustness guarantees.
PhD thesis,
University of California, Los Angeles
(2020)
\end{botherref}
\endbibitem

\bibitem[\protect\citeauthoryear{Xu et~al.}{2020}]{xu2020automatic}
\begin{barticle}
\bauthor{\bsnm{Xu}, \binits{K.}},
\bauthor{\bsnm{Shi}, \binits{Z.}},
\bauthor{\bsnm{Zhang}, \binits{H.}},
\bauthor{\bsnm{Wang}, \binits{Y.}},
\bauthor{\bsnm{Chang}, \binits{K.-W.}},
\bauthor{\bsnm{Huang}, \binits{M.}},
\bauthor{\bsnm{Kailkhura}, \binits{B.}},
\bauthor{\bsnm{Lin}, \binits{X.}},
\bauthor{\bsnm{Hsieh}, \binits{C.-J.}}:
\batitle{Automatic perturbation analysis for scalable certified robustness and beyond}.
\bjtitle{Advances in Neural Information Processing Systems}
\bvolume{33},
\bfpage{1129}--\blpage{1141}
(\byear{2020})
\end{barticle}
\endbibitem

\bibitem[\protect\citeauthoryear{Yang et~al.}{2023}]{yang2023provable}
\begin{bchapter}
\bauthor{\bsnm{Yang}, \binits{R.}},
\bauthor{\bsnm{Laurel}, \binits{J.}},
\bauthor{\bsnm{Misailovic}, \binits{S.}},
\bauthor{\bsnm{Singh}, \binits{G.}}:
\bctitle{Provable defense against geometric transformations}.
In: \bbtitle{Proceeding of the Eleventh International Conference on Learning Representations},
pp. \bfpage{1}--\blpage{19}
(\byear{2023})
\end{bchapter}
\endbibitem

\bibitem[\protect\citeauthoryear{Dong et~al.}{2024}]{dong2024robust}
\begin{bchapter}
\bauthor{\bsnm{Dong}, \binits{J.}},
\bauthor{\bsnm{Koniusz}, \binits{P.}},
\bauthor{\bsnm{Chen}, \binits{J.}},
\bauthor{\bsnm{Wang}, \binits{Z.J.}},
\bauthor{\bsnm{Ong}, \binits{Y.-S.}}:
\bctitle{Robust distillation via untargeted and targeted intermediate adversarial samples}.
In: \bbtitle{Proceedings of the IEEE/CVF Conference on Computer Vision and Pattern Recognition (CVPR)},
pp. \bfpage{28432}--\blpage{28442}
(\byear{2024})
\end{bchapter}
\endbibitem

\bibitem[\protect\citeauthoryear{Liang and Samavi}{2023}]{liang2023advanced}
\begin{barticle}
\bauthor{\bsnm{Liang}, \binits{Y.}},
\bauthor{\bsnm{Samavi}, \binits{R.}}:
\batitle{Advanced defensive distillation with ensemble voting and noisy logits}.
\bjtitle{Applied Intelligence}
\bvolume{53}(\bissue{3}),
\bfpage{3069}--\blpage{3094}
(\byear{2023})
\end{barticle}
\endbibitem

\bibitem[\protect\citeauthoryear{Kuang et~al.}{2023}]{kuang2023improving}
\begin{barticle}
\bauthor{\bsnm{Kuang}, \binits{H.}},
\bauthor{\bsnm{Liu}, \binits{H.}},
\bauthor{\bsnm{Wu}, \binits{Y.}},
\bauthor{\bsnm{Satoh}, \binits{S.}},
\bauthor{\bsnm{Ji}, \binits{R.}}:
\batitle{Improving adversarial robustness via information bottleneck distillation}.
\bjtitle{Advances in Neural Information Processing Systems}
\bvolume{36},
\bfpage{10796}--\blpage{10813}
(\byear{2023})
\end{barticle}
\endbibitem

\bibitem[\protect\citeauthoryear{Zhang et~al.}{2022}]{zhang2022towards}
\begin{barticle}
\bauthor{\bsnm{Zhang}, \binits{H.}},
\bauthor{\bsnm{Fu}, \binits{Z.}},
\bauthor{\bsnm{Li}, \binits{G.}},
\bauthor{\bsnm{Ma}, \binits{L.}},
\bauthor{\bsnm{Zhao}, \binits{Z.}},
\bauthor{\bsnm{Yang}, \binits{H.}},
\bauthor{\bsnm{Sun}, \binits{Y.}},
\bauthor{\bsnm{Liu}, \binits{Y.}},
\bauthor{\bsnm{Jin}, \binits{Z.}}:
\batitle{Towards robustness of deep program processing models—detection, estimation, and enhancement}.
\bjtitle{ACM Transactions on Software Engineering and Methodology (TOSEM)}
\bvolume{31}(\bissue{3}),
\bfpage{1}--\blpage{40}
(\byear{2022})
\end{barticle}
\endbibitem

\bibitem[\protect\citeauthoryear{Fang et~al.}{2022}]{fang2022enhanced}
\begin{barticle}
\bauthor{\bsnm{Fang}, \binits{Y.}},
\bauthor{\bsnm{Xiao}, \binits{S.}},
\bauthor{\bsnm{Zhou}, \binits{M.}},
\bauthor{\bsnm{Cai}, \binits{S.}},
\bauthor{\bsnm{Zhang}, \binits{Z.}}:
\batitle{Enhanced task attention with adversarial learning for dynamic multi-task cnn}.
\bjtitle{Pattern Recognition}
\bvolume{128},
\bfpage{108672}
(\byear{2022})
\end{barticle}
\endbibitem

\bibitem[\protect\citeauthoryear{Leite and Xiao}{2020}]{leite2020improving}
\begin{barticle}
\bauthor{\bsnm{Leite}, \binits{C.F.S.}},
\bauthor{\bsnm{Xiao}, \binits{Y.}}:
\batitle{Improving cross-subject activity recognition via adversarial learning}.
\bjtitle{IEEE Access}
\bvolume{8},
\bfpage{90542}--\blpage{90554}
(\byear{2020})
\end{barticle}
\endbibitem

\bibitem[\protect\citeauthoryear{Gupta}{2022}]{gupta2022non}
\begin{barticle}
\bauthor{\bsnm{Gupta}, \binits{S.}}:
\batitle{Non-functional requirements elicitation for edge computing}.
\bjtitle{Internet of Things}
\bvolume{18},
\bfpage{100503}
(\byear{2022})
\end{barticle}
\endbibitem

\bibitem[\protect\citeauthoryear{Moosavi-Dezfooli et~al.}{2016}]{moosavi2016deepfool}
\begin{bchapter}
\bauthor{\bsnm{Moosavi-Dezfooli}, \binits{S.-M.}},
\bauthor{\bsnm{Fawzi}, \binits{A.}},
\bauthor{\bsnm{Frossard}, \binits{P.}}:
\bctitle{Deepfool: a simple and accurate method to fool deep neural networks}.
In: \bbtitle{Proceedings of the IEEE Conference on Computer Vision and Pattern Recognition},
pp. \bfpage{2574}--\blpage{2582}
(\byear{2016})
\end{bchapter}
\endbibitem

\bibitem[\protect\citeauthoryear{Kurakin et~al.}{2018}]{kurakin2018adversarial}
\begin{bchapter}
\bauthor{\bsnm{Kurakin}, \binits{A.}},
\bauthor{\bsnm{Goodfellow}, \binits{I.J.}},
\bauthor{\bsnm{Bengio}, \binits{S.}}:
\bctitle{Adversarial examples in the physical world}.
In: \bbtitle{Proceedings of the Artificial Intelligence Safety and Security},
pp. \bfpage{99}--\blpage{112}.
\bpublisher{Chapman and Hall/CRC}, \blocation{???}
(\byear{2018})
\end{bchapter}
\endbibitem

\bibitem[\protect\citeauthoryear{Chen et~al.}{2018}]{chen2018ead}
\begin{bchapter}
\bauthor{\bsnm{Chen}, \binits{P.-Y.}},
\bauthor{\bsnm{Sharma}, \binits{Y.}},
\bauthor{\bsnm{Zhang}, \binits{H.}},
\bauthor{\bsnm{Yi}, \binits{J.}},
\bauthor{\bsnm{Hsieh}, \binits{C.-J.}}:
\bctitle{Ead: elastic-net attacks to deep neural networks via adversarial examples}.
In: \bbtitle{Proceedings of the AAAI Conference on Artificial Intelligence},
vol. \bseriesno{32},
pp. \bfpage{1}--\blpage{10}
(\byear{2018})
\end{bchapter}
\endbibitem

\bibitem[\protect\citeauthoryear{Uesato et~al.}{2018}]{uesato2018adversarial}
\begin{bchapter}
\bauthor{\bsnm{Uesato}, \binits{J.}},
\bauthor{\bsnm{O'Donoghue}, \binits{B.}},
\bauthor{\bsnm{Kohli}, \binits{P.}},
\bauthor{\bsnm{Oord}, \binits{A.}}:
\bctitle{Adversarial risk and the dangers of evaluating against weak attacks}.
In: \bbtitle{Proceedings of the 35th International Conference on Machine Learning},
vol. \bseriesno{80},
pp. \bfpage{5025}--\blpage{5034}.
\bpublisher{PMLR}, \blocation{???}
(\byear{2018})
\end{bchapter}
\endbibitem

\bibitem[\protect\citeauthoryear{Papernot et~al.}{2016}]{papernot2016limitations}
\begin{bchapter}
\bauthor{\bsnm{Papernot}, \binits{N.}},
\bauthor{\bsnm{McDaniel}, \binits{P.}},
\bauthor{\bsnm{Jha}, \binits{S.}},
\bauthor{\bsnm{Fredrikson}, \binits{M.}},
\bauthor{\bsnm{Celik}, \binits{Z.B.}},
\bauthor{\bsnm{Swami}, \binits{A.}}:
\bctitle{The limitations of deep learning in adversarial settings}.
In: \bbtitle{Proceedings of the IEEE European Symposium on Security and Privacy (EuroS\&P)},
pp. \bfpage{372}--\blpage{387}
(\byear{2016}).
\bcomment{IEEE}
\end{bchapter}
\endbibitem

\bibitem[\protect\citeauthoryear{Zhang et~al.}{2018}]{zhang2018efficient}
\begin{botherref}
\oauthor{\bsnm{Zhang}, \binits{H.}},
\oauthor{\bsnm{Weng}, \binits{T.-W.}},
\oauthor{\bsnm{Chen}, \binits{P.-Y.}},
\oauthor{\bsnm{Hsieh}, \binits{C.-J.}},
\oauthor{\bsnm{Daniel}, \binits{L.}}:
Efficient neural network robustness certification with general activation functions.
Advances in neural information processing systems
\textbf{31}
(2018)
\end{botherref}
\endbibitem

\bibitem[\protect\citeauthoryear{Gowal et~al.}{2019}]{gowal2019scalable}
\begin{bchapter}
\bauthor{\bsnm{Gowal}, \binits{S.}},
\bauthor{\bsnm{Dvijotham}, \binits{K.D.}},
\bauthor{\bsnm{Stanforth}, \binits{R.}},
\bauthor{\bsnm{Bunel}, \binits{R.}},
\bauthor{\bsnm{Qin}, \binits{C.}},
\bauthor{\bsnm{Uesato}, \binits{J.}},
\bauthor{\bsnm{Arandjelovic}, \binits{R.}},
\bauthor{\bsnm{Mann}, \binits{T.}},
\bauthor{\bsnm{Kohli}, \binits{P.}}:
\bctitle{Scalable verified training for provably robust image classification}.
In: \bbtitle{Proceedings of the IEEE/CVF International Conference on Computer Vision},
pp. \bfpage{4842}--\blpage{4851}
(\byear{2019})
\end{bchapter}
\endbibitem

\bibitem[\protect\citeauthoryear{Reiss and Stricker}{2012}]{reiss2012introducing}
\begin{bchapter}
\bauthor{\bsnm{Reiss}, \binits{A.}},
\bauthor{\bsnm{Stricker}, \binits{D.}}:
\bctitle{Introducing a new benchmarked dataset for activity monitoring}.
In: \bbtitle{Proceeding of the 16th International Symposium on Wearable Computers},
pp. \bfpage{108}--\blpage{109}
(\byear{2012}).
\bcomment{IEEE}
\end{bchapter}
\endbibitem

\bibitem[\protect\citeauthoryear{Liu et~al.}{2015}]{liu2015deep}
\begin{bchapter}
\bauthor{\bsnm{Liu}, \binits{Z.}},
\bauthor{\bsnm{Luo}, \binits{P.}},
\bauthor{\bsnm{Wang}, \binits{X.}},
\bauthor{\bsnm{Tang}, \binits{X.}}:
\bctitle{Deep learning face attributes in the wild}.
In: \bbtitle{Proceedings of the IEEE International Conference on Computer Vision},
pp. \bfpage{3730}--\blpage{3738}
(\byear{2015})
\end{bchapter}
\endbibitem

\bibitem[\protect\citeauthoryear{Lin et~al.}{2019}]{lin2019improving}
\begin{barticle}
\bauthor{\bsnm{Lin}, \binits{Y.}},
\bauthor{\bsnm{Zheng}, \binits{L.}},
\bauthor{\bsnm{Zheng}, \binits{Z.}},
\bauthor{\bsnm{Wu}, \binits{Y.}},
\bauthor{\bsnm{Hu}, \binits{Z.}},
\bauthor{\bsnm{Yan}, \binits{C.}},
\bauthor{\bsnm{Yang}, \binits{Y.}}:
\batitle{Improving person re-identification by attribute and identity learning}.
\bjtitle{Pattern recognition}
\bvolume{95},
\bfpage{151}--\blpage{161}
(\byear{2019})
\end{barticle}
\endbibitem

\bibitem[\protect\citeauthoryear{Li et~al.}{2017}]{li2017reliable}
\begin{bchapter}
\bauthor{\bsnm{Li}, \binits{S.}},
\bauthor{\bsnm{Deng}, \binits{W.}},
\bauthor{\bsnm{Du}, \binits{J.}}:
\bctitle{Reliable crowdsourcing and deep locality-preserving learning for expression recognition in the wild}.
In: \bbtitle{Proceedings of the IEEE Conference on Computer Vision and Pattern Recognition},
pp. \bfpage{2852}--\blpage{2861}
(\byear{2017})
\end{bchapter}
\endbibitem

\bibitem[\protect\citeauthoryear{Luo et~al.}{2022}]{luo2022frequency}
\begin{bchapter}
\bauthor{\bsnm{Luo}, \binits{C.}},
\bauthor{\bsnm{Lin}, \binits{Q.}},
\bauthor{\bsnm{Xie}, \binits{W.}},
\bauthor{\bsnm{Wu}, \binits{B.}},
\bauthor{\bsnm{Xie}, \binits{J.}},
\bauthor{\bsnm{Shen}, \binits{L.}}:
\bctitle{Frequency-driven imperceptible adversarial attack on semantic similarity}.
In: \bbtitle{Proceedings of the IEEE/CVF Conference on Computer Vision and Pattern Recognition},
pp. \bfpage{15315}--\blpage{15324}
(\byear{2022})
\end{bchapter}
\endbibitem

\bibitem[\protect\citeauthoryear{Xiao et~al.}{2018}]{xiao2018spatially}
\begin{bchapter}
\bauthor{\bsnm{Xiao}, \binits{C.}},
\bauthor{\bsnm{Zhu}, \binits{J.-Y.}},
\bauthor{\bsnm{Li}, \binits{B.}},
\bauthor{\bsnm{He}, \binits{W.}},
\bauthor{\bsnm{Liu}, \binits{M.}},
\bauthor{\bsnm{Song}, \binits{D.}}:
\bctitle{Spatially transformed adversarial examples}.
In: \bbtitle{Proceedings of the International Conference on Learning Representations}
(\byear{2018})
\end{bchapter}
\endbibitem

\bibitem[\protect\citeauthoryear{Mao et~al.}{2021}]{mao2021composite}
\begin{bchapter}
\bauthor{\bsnm{Mao}, \binits{X.}},
\bauthor{\bsnm{Chen}, \binits{Y.}},
\bauthor{\bsnm{Wang}, \binits{S.}},
\bauthor{\bsnm{Su}, \binits{H.}},
\bauthor{\bsnm{He}, \binits{Y.}},
\bauthor{\bsnm{Xue}, \binits{H.}}:
\bctitle{Composite adversarial attacks}.
In: \bbtitle{Proceedings of the AAAI Conference on Artificial Intelligence},
vol. \bseriesno{35},
pp. \bfpage{8884}--\blpage{8892}
(\byear{2021})
\end{bchapter}
\endbibitem

\bibitem[\protect\citeauthoryear{Luo and Kong}{2024}]{luo2024enhancing}
\begin{bchapter}
\bauthor{\bsnm{Luo}, \binits{J.}},
\bauthor{\bsnm{Kong}, \binits{L.}}:
\bctitle{On enhancing adversarial robustness of large pre-trained vision-language models}.
In: \bbtitle{Proceedings of the 2024 8th International Conference on Computer Science and Artificial Intelligence},
pp. \bfpage{212}--\blpage{220}
(\byear{2024})
\end{bchapter}
\endbibitem

\bibitem[\protect\citeauthoryear{Noack et~al.}{2021}]{noack2021empirical}
\begin{barticle}
\bauthor{\bsnm{Noack}, \binits{A.}},
\bauthor{\bsnm{Ahern}, \binits{I.}},
\bauthor{\bsnm{Dou}, \binits{D.}},
\bauthor{\bsnm{Li}, \binits{B.}}:
\batitle{An empirical study on the relation between network interpretability and adversarial robustness}.
\bjtitle{SN Computer Science}
\bvolume{2},
\bfpage{1}--\blpage{13}
(\byear{2021})
\end{barticle}
\endbibitem

\bibitem[\protect\citeauthoryear{H{\"u}llermeier and Waegeman}{2021}]{hullermeier2021aleatoric}
\begin{barticle}
\bauthor{\bsnm{H{\"u}llermeier}, \binits{E.}},
\bauthor{\bsnm{Waegeman}, \binits{W.}}:
\batitle{Aleatoric and epistemic uncertainty in machine learning: An introduction to concepts and methods}.
\bjtitle{Machine Learning}
\bvolume{110},
\bfpage{457}--\blpage{506}
(\byear{2021})
\end{barticle}
\endbibitem

\bibitem[\protect\citeauthoryear{Li et~al.}{2020}]{li2020learnable}
\begin{barticle}
\bauthor{\bsnm{Li}, \binits{J.}},
\bauthor{\bsnm{Liu}, \binits{H.}},
\bauthor{\bsnm{Tao}, \binits{Z.}},
\bauthor{\bsnm{Zhao}, \binits{H.}},
\bauthor{\bsnm{Fu}, \binits{Y.}}:
\batitle{Learnable subspace clustering}.
\bjtitle{IEEE Transactions on Neural Networks and Learning Systems}
\bvolume{33}(\bissue{3}),
\bfpage{1119}--\blpage{1133}
(\byear{2020})
\end{barticle}
\endbibitem

\bibitem[\protect\citeauthoryear{Guo et~al.}{2023}]{guo2023comprehensive}
\begin{botherref}
\oauthor{\bsnm{Guo}, \binits{J.}},
\oauthor{\bsnm{Bao}, \binits{W.}},
\oauthor{\bsnm{Wang}, \binits{J.}},
\oauthor{\bsnm{Ma}, \binits{Y.}},
\oauthor{\bsnm{Gao}, \binits{X.}},
\oauthor{\bsnm{Xiao}, \binits{G.}},
\oauthor{\bsnm{Liu}, \binits{A.}},
\oauthor{\bsnm{Dong}, \binits{J.}},
\oauthor{\bsnm{Liu}, \binits{X.}},
\oauthor{\bsnm{Wu}, \binits{W.}}:
A comprehensive evaluation framework for deep model robustness.
Pattern Recognition,
109308
(2023)
\end{botherref}
\endbibitem

\bibitem[\protect\citeauthoryear{Feng et~al.}{2023}]{feng2023dynamic}
\begin{bchapter}
\bauthor{\bsnm{Feng}, \binits{W.}},
\bauthor{\bsnm{Xu}, \binits{N.}},
\bauthor{\bsnm{Zhang}, \binits{T.}},
\bauthor{\bsnm{Zhang}, \binits{Y.}}:
\bctitle{Dynamic generative targeted attacks with pattern injection}.
In: \bbtitle{Proceedings of the IEEE/CVF Conference on Computer Vision and Pattern Recognition},
pp. \bfpage{16404}--\blpage{16414}
(\byear{2023})
\end{bchapter}
\endbibitem

\bibitem[\protect\citeauthoryear{Kienitz et~al.}{2022}]{kienitz2022comparing}
\begin{bchapter}
\bauthor{\bsnm{Kienitz}, \binits{D.}},
\bauthor{\bsnm{Komendantskaya}, \binits{E.}},
\bauthor{\bsnm{A~Lones}, \binits{M.}}:
\bctitle{Comparing complexities of decision boundaries for robust training: A universal approach}.
In: \bbtitle{Proceedings of the Asian Conference on Computer Vision},
pp. \bfpage{4495}--\blpage{4513}
(\byear{2022})
\end{bchapter}
\endbibitem

\bibitem[\protect\citeauthoryear{Chen et~al.}{2021}]{chen2021local}
\begin{bchapter}
\bauthor{\bsnm{Chen}, \binits{S.}},
\bauthor{\bsnm{Yao}, \binits{T.}},
\bauthor{\bsnm{Chen}, \binits{Y.}},
\bauthor{\bsnm{Ding}, \binits{S.}},
\bauthor{\bsnm{Li}, \binits{J.}},
\bauthor{\bsnm{Ji}, \binits{R.}}:
\bctitle{Local relation learning for face forgery detection}.
In: \bbtitle{Proceedings of the AAAI Conference on Artificial Intelligence},
vol. \bseriesno{35},
pp. \bfpage{1081}--\blpage{1088}
(\byear{2021})
\end{bchapter}
\endbibitem

\bibitem[\protect\citeauthoryear{Li et~al.}{2020}]{li2020adversarial}
\begin{barticle}
\bauthor{\bsnm{Li}, \binits{F.}},
\bauthor{\bsnm{Lai}, \binits{L.}},
\bauthor{\bsnm{Cui}, \binits{S.}}:
\batitle{On the adversarial robustness of subspace learning}.
\bjtitle{IEEE Transactions on Signal Processing}
\bvolume{68},
\bfpage{1470}--\blpage{1483}
(\byear{2020})
\end{barticle}
\endbibitem

\bibitem[\protect\citeauthoryear{Jiang and Ge}{2022}]{jiang2022attacks}
\begin{barticle}
\bauthor{\bsnm{Jiang}, \binits{X.}},
\bauthor{\bsnm{Ge}, \binits{Z.}}:
\batitle{Attacks on data-driven process monitoring systems: Subspace transfer networks}.
\bjtitle{IEEE Transactions on Artificial Intelligence}
\bvolume{3}(\bissue{3}),
\bfpage{470}--\blpage{484}
(\byear{2022})
\end{barticle}
\endbibitem

\bibitem[\protect\citeauthoryear{Fu et~al.}{2023}]{fu2023multi}
\begin{barticle}
\bauthor{\bsnm{Fu}, \binits{G.}},
\bauthor{\bsnm{Zhang}, \binits{Z.}},
\bauthor{\bsnm{Le}, \binits{W.}},
\bauthor{\bsnm{Li}, \binits{J.}},
\bauthor{\bsnm{Zhu}, \binits{Q.}},
\bauthor{\bsnm{Niu}, \binits{F.}},
\bauthor{\bsnm{Chen}, \binits{H.}},
\bauthor{\bsnm{Sun}, \binits{F.}},
\bauthor{\bsnm{Shen}, \binits{Y.}}:
\batitle{A multi-scale pooling convolutional neural network for accurate steel surface defects classification}.
\bjtitle{Frontiers in Neurorobotics}
\bvolume{17},
\bfpage{1096083}
(\byear{2023})
\end{barticle}
\endbibitem

\bibitem[\protect\citeauthoryear{Karras et~al.}{2020}]{karras2020analyzing}
\begin{bchapter}
\bauthor{\bsnm{Karras}, \binits{T.}},
\bauthor{\bsnm{Laine}, \binits{S.}},
\bauthor{\bsnm{Aittala}, \binits{M.}},
\bauthor{\bsnm{Hellsten}, \binits{J.}},
\bauthor{\bsnm{Lehtinen}, \binits{J.}},
\bauthor{\bsnm{Aila}, \binits{T.}}:
\bctitle{Analyzing and improving the image quality of {StyleGAN}}.
In: \bbtitle{Proceedings of the IEEE/CVF Conference on Computer Vision and Pattern Recognition},
pp. \bfpage{8110}--\blpage{8119}
(\byear{2020})
\end{bchapter}
\endbibitem

\bibitem[\protect\citeauthoryear{Zhou et~al.}{2022a}]{zhou2022adversarial}
\begin{barticle}
\bauthor{\bsnm{Zhou}, \binits{S.}},
\bauthor{\bsnm{Liu}, \binits{C.}},
\bauthor{\bsnm{Ye}, \binits{D.}},
\bauthor{\bsnm{Zhu}, \binits{T.}},
\bauthor{\bsnm{Zhou}, \binits{W.}},
\bauthor{\bsnm{Yu}, \binits{P.S.}}:
\batitle{Adversarial attacks and defenses in deep learning: From a perspective of cybersecurity}.
\bjtitle{ACM Computing Surveys}
\bvolume{55}(\bissue{8}),
\bfpage{1}--\blpage{39}
(\byear{2022})
\end{barticle}
\endbibitem

\bibitem[\protect\citeauthoryear{Zhou et~al.}{2022b}]{zhou2022understanding}
\begin{bchapter}
\bauthor{\bsnm{Zhou}, \binits{D.}},
\bauthor{\bsnm{Yu}, \binits{Z.}},
\bauthor{\bsnm{Xie}, \binits{E.}},
\bauthor{\bsnm{Xiao}, \binits{C.}},
\bauthor{\bsnm{Anandkumar}, \binits{A.}},
\bauthor{\bsnm{Feng}, \binits{J.}},
\bauthor{\bsnm{Alvarez}, \binits{J.M.}}:
\bctitle{Understanding the robustness in vision transformers}.
In: \bbtitle{Proceedings of the International Conference on Machine Learning},
pp. \bfpage{27378}--\blpage{27394}
(\byear{2022}).
\bcomment{PMLR}
\end{bchapter}
\endbibitem

\bibitem[\protect\citeauthoryear{Weng et~al.}{2018}]{weng2018towards}
\begin{bchapter}
\bauthor{\bsnm{Weng}, \binits{L.}},
\bauthor{\bsnm{Zhang}, \binits{H.}},
\bauthor{\bsnm{Chen}, \binits{H.}},
\bauthor{\bsnm{Song}, \binits{Z.}},
\bauthor{\bsnm{Hsieh}, \binits{C.-J.}},
\bauthor{\bsnm{Daniel}, \binits{L.}},
\bauthor{\bsnm{Boning}, \binits{D.}},
\bauthor{\bsnm{Dhillon}, \binits{I.}}:
\bctitle{Towards fast computation of certified robustness for relu networks}.
In: \bbtitle{Proceedings of the International Conference on Machine Learning},
pp. \bfpage{5276}--\blpage{5285}
(\byear{2018}).
\bcomment{PMLR}
\end{bchapter}
\endbibitem

\end{thebibliography}

\end{document}